\documentclass[10pt,twocolumn,letterpaper]{article}

\usepackage{cvpr}              

\usepackage[dvipsnames]{xcolor}

\definecolor{cvprblue}{rgb}{0.21,0.49,0.74}
\usepackage[pagebackref,breaklinks,colorlinks,citecolor=cvprblue]{hyperref}

\def\paperID{2416} 
\def\confName{CVPR}
\def\confYear{2024}

\title{Gradient-based Parameter Selection for Efficient Fine-Tuning}

\newcommand*{\affaddr}[1]{#1} 
\newcommand*{\affmark}[1][*]{\textsuperscript{#1}}

\author{
Zhi Zhang\affmark[1,2]\footnotemark[1], Qizhe Zhang\affmark[2]\footnotemark[1], Zijun Gao\affmark[2], Renrui Zhang\affmark[3,4], Ekaterina Shutova\affmark[1], Shiji Zhou\affmark[5], Shanghang Zhang\affmark[2]\footnotemark[2]\\
\affaddr{\affmark[1]ILLC, University of Amsterdam}, 
\\
\affaddr{\affmark[2]National Key Laboratory for Multimedia Information Processing, \\School of Computer Science, Peking University},
\\
\affaddr{\affmark[3]MMLAB, CUHK}, \affaddr{\affmark[4]Shanghai AI Laboratory}
\\
\affaddr{\affmark[5]Department of Automation, Tsinghua University},
\\
{\tt\small \{z.zhang, e.shutova\}@uva.nl, \{theia, shanghang\}@pku.edu.cn} 
}

\begin{document}
\maketitle

\renewcommand{\thefootnote}{\fnsymbol{footnote}} 
\footnotetext[1]{Contributed equally.} 
\footnotetext[2]{Corresponding author.} 

\begin{abstract}
With the growing size of pre-trained models, full fine-tuning and storing all the parameters for various downstream tasks is costly and infeasible. In this paper, we propose a new parameter-efficient fine-tuning method, \textbf{G}radient-based \textbf{P}arameter \textbf{S}election (GPS),  demonstrating that only tuning a few selected parameters from the pre-trained model while keeping the remainder of the model frozen can generate similar or better performance compared with the full model fine-tuning method. Different from the existing popular and state-of-the-art parameter-efficient fine-tuning approaches, our method does not introduce any additional parameters and computational costs during both the training and inference stages. Another advantage is the model-agnostic and non-destructive property, which eliminates the need for any other design specific to a particular model. Compared with the full fine-tuning, GPS achieves 3.33\% (91.78\% vs. 88.45\%, FGVC) and 9.61\% (73.1\% vs. 65.57\%, VTAB) improvement of the accuracy with tuning only 0.36\% parameters of the pre-trained model on average over 24 image classification tasks;  it also demonstrates a significant improvement of 17\% and 16.8\% in mDice and mIoU, respectively, on medical image segmentation task. Moreover, GPS achieves state-of-the-art performance compared with existing PEFT methods. The code will be available in \url{https://github.com/FightingFighting/GPS.git}.
\end{abstract}    
\section{Introduction}
\label{sec:intro}

\begin{figure}[!t]
\centering
\includegraphics[width=1\linewidth]{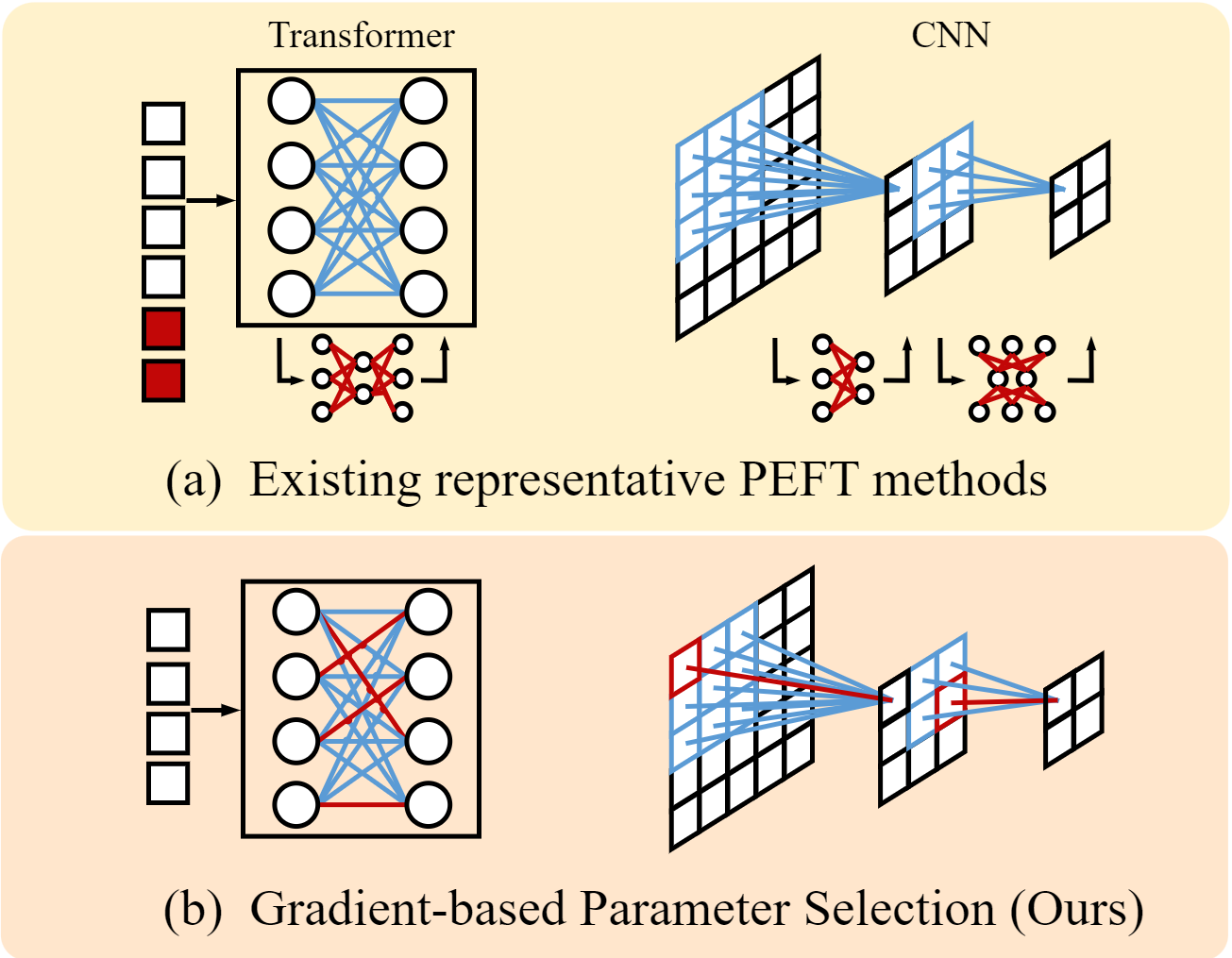}
\caption{Comparison between our GPS and other PEFT methods. (a) Exiting popular methods introduce extra parameters for tuning downstream tasks, which might need a special design for diverse architectures, such as appending prompt into the input token in Transformer or inserting different modules into different layers (b) Our approach avoids the introduction of additional parameters and solely fine-tunes the selected parameters from the model, employing a unified gradient-based parameter selection method across diverse architectural variations, e.g. Transformer and CNN.}
\label{fig: differences with other methods}
\vspace{-0.1cm}
\end{figure}
The pre-training and fine-tuning pipeline has become a common paradigm for adapting large models pre-trained on substantial amounts of data to downstream tasks with fewer training samples. However, fine-tuning all the parameters in the model is memory-intensive and data-inefficient, which is costly and infeasible for multiple downstream tasks given a large-scale model~\cite{lian2022scaling, jia2022visual, houlsby2019parameter}. To tackle this issue, parameter-efficient fine-tuning (PEFT) methods have been proposed with the aim of tuning a minimal number of parameters to fit downstream tasks while keeping most of the parameters frozen. Another benefit of PEFT is that tuning a smaller set of parameters reduces the complexity of optimization and alleviates overfitting concerns when adapting large pre-trained models to downstream tasks with limited data, resulting in comparable or even superior performance compared to full fine-tuning~\cite{jia2022visual}. 
Inspired by the success of PEFT in NLP
\cite{2023Parameter,hu2022sparse, su2021transferability,hu2021lora,li2021prefix, He2021TowardsAU}, several methods have been introduced to vision tasks, such as Adapter~\cite{houlsby2019parameter} and Visual Prompt Tuning (VPT)~\cite{jia2022visual} introducing extra learnable parameters into the backbone and the input space of the pre-trained model respectively.
SSF, another representative method, transforms features across layers of the pre-trained model using extra learnable layers~\cite{lian2022scaling}. 


However, these methods introduce additional parameters into the pre-trained model and disrupt its original architecture, leading to increased computational costs during training and/or inference stages. Furthermore, these approaches lack generalizability across various model architectures. Specifically, different models are equipped with distinct components (layers), such as MLPs, activation functions, and self-attention layers. These methods need to determine the optimal locations for inserting extra parameters between different layers; moreover, certain transformer-based techniques cannot be directly applied to convolution-based methods like VPT. Therefore, these methods exhibit limited compatibility with diverse architectures.

\begin{figure}[!t]
\centering
\subfigure[VTAB-1K] {\includegraphics[width=0.495\linewidth]{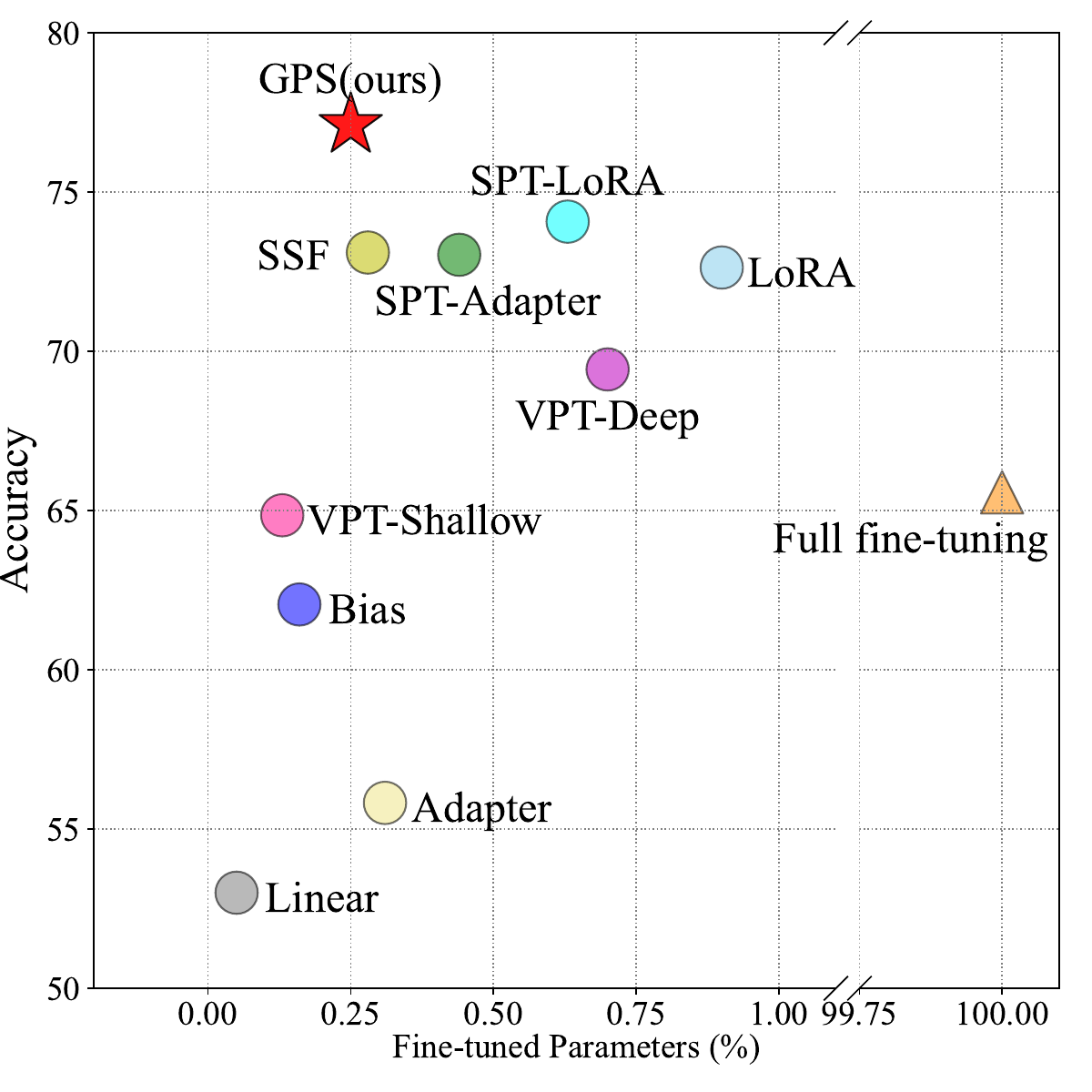}}
\subfigure[FGVC] {\includegraphics[width=0.495\linewidth]{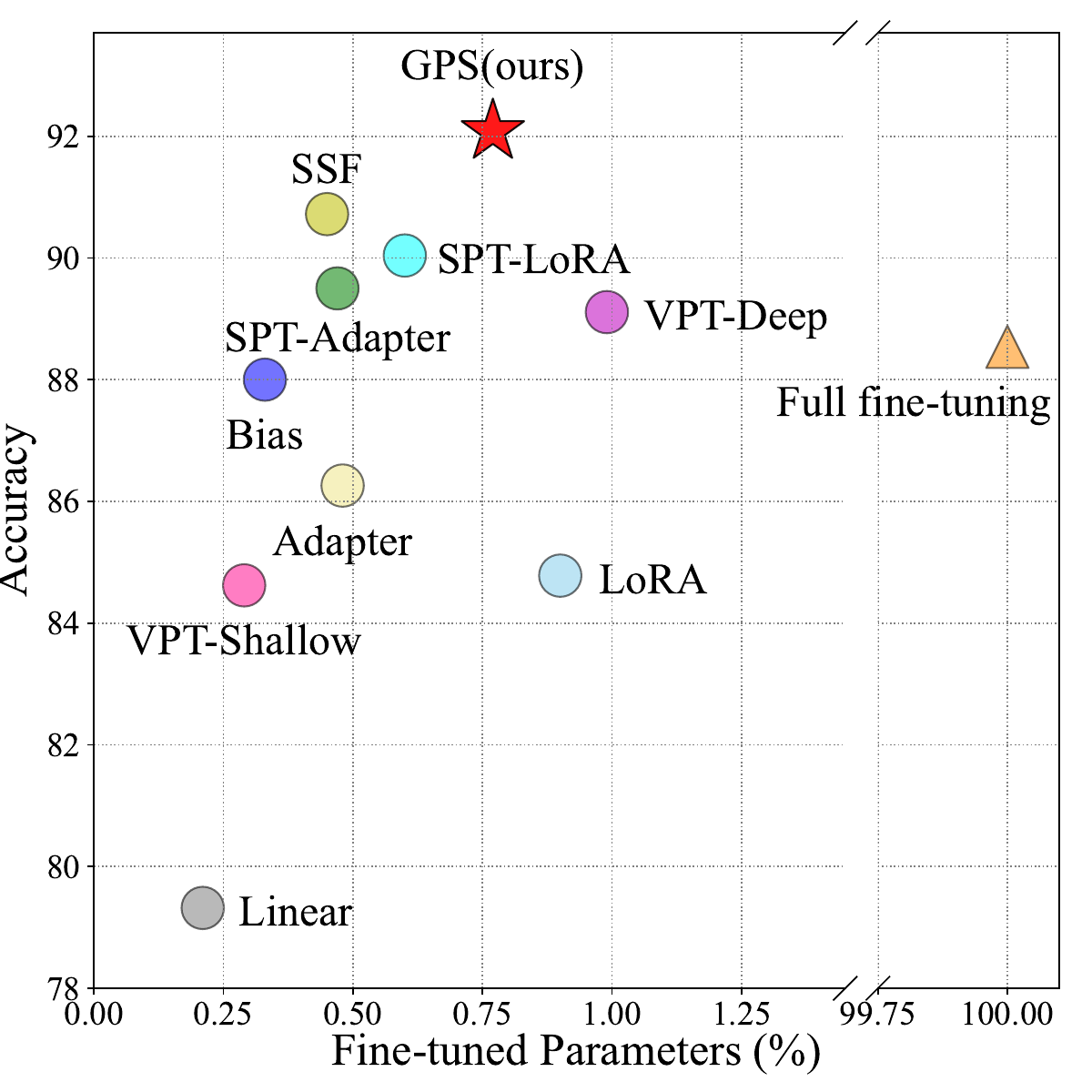}}
\caption{Performance comparisons of 11 fine-tuning methods with a pre-trained ViT-B/16 model on the VTAB-1k (a) and FGVC (b) benchmarks. Our GPS (red stars) achieves state-of-the-art performance on both benchmarks with only 0.25\% and 0.77\% average trainable parameters respectively.}
\label{fig:comparison}
\end{figure}

To tackle the above issues, we propose a non-destructive network architecture and model-agnostic PEFT approach, which introduces no extra parameters during both training and test stages and provides a unified solution for various architectures. 
We select a small number of essential parameters from the pre-trained model and only fine-tune these parameters for the downstream tasks. 
To select them, we propose a fine-grained Gradient-based Parameter Selection (GPS) method. 
For each neuron in the network, we choose top-K of its input connections (weights or parameters) with the highest gradient value, resulting in a small proportion of the parameters in the model being selected.

\begin{table}
    \centering
        \setlength{\tabcolsep}{2.5pt}
        \resizebox{1\linewidth}{!}{
            \begin{tabular}{@{}ccccccc@{}}
            \toprule
            \multirow{2}{*}{Method} & Mean & Params. & Model  & No extra  & No extra  & Task  \\
             &Acc. &  (\%)   & Agnostic & Train param.  & Infer params. & Adaptive  \\
             \midrule
            Full~\cite{jia2022visual} &  70.36  &  100  & \multicolumn{1}{c}{\ding{51}} & \multicolumn{1}{c}{\ding{51}} &  \multicolumn{1}{c}{\ding{51}} & {\ding{55}}\\
            \midrule
            Linear~\cite{jia2022visual} & 58.48   &  0.08  & \multicolumn{1}{c}{\ding{51}} & \multicolumn{1}{c}{\ding{51}} &  \multicolumn{1}{c}{\ding{51}} & {\ding{55}}                   \\ 
            Bias~\cite{zaken2021bitfit} & 67.54   &  0.20  & \multicolumn{1}{c}{\ding{51}} & \multicolumn{1}{c}{\ding{51}} &  \multicolumn{1}{c}{\ding{51}}  & {\ding{55}}                   \\ 
            \midrule
            Adapter~\cite{houlsby2019parameter}        & 60.04 & 0.35 & \multicolumn{1}{c}{\ding{55}} & \multicolumn{1}{c}{\ding{55}}& \multicolumn{1}{c}{\ding{55}}  & {\ding{55}}                    \\
            VPT~\cite{jia2022visual}            & 73.53 & 0.76                & \multicolumn{1}{c}{\ding{55}} &\multicolumn{1}{c}{\ding{55}} & \multicolumn{1}{c}{\ding{55}}   & {\ding{55}}           \\
            LoRA~\cite{hu2021lora}        & 75.16 & 0.90 & \multicolumn{1}{c}{\ding{55}} & \multicolumn{1}{c}{\ding{55}}& \multicolumn{1}{c}{\ding{51}}  & {\ding{55}}                    \\
            SSF~\cite{lian2022scaling}            & 76.77 & 0.32 & \multicolumn{1}{c}{\ding{55}}  & \multicolumn{1}{c}{\ding{55}}  & \multicolumn{1}{c}{\ding{51}}    & {\ding{55}}
                 \\ \midrule
            GPS (ours)     &78.64   & 0.36  &  \multicolumn{1}{c}{\ding{51}} & \multicolumn{1}{c}{\ding{51}}  & \multicolumn{1}{c}{\ding{51}}  & {\ding{51}}                  \\ \bottomrule
            \end{tabular}
        }
         \caption{Comparison between different fine-tuning methods. The ViT-B/16 model accuracy over all 24 tasks in FGVA and VTAB fine-tuned on ViT-B/16 model and the number of tunable parameters are shown in columns Acc. and Params. (\%). 
         }
        \label{table: Comparison between different fine-tuning methods}
\end{table}

Such design offers five-fold benefits: 
\romannumeral1) The pre-trained model can efficiently tackle downstream tasks because the gradient direction indicates the fastest loss function changes and highest change rate, facilitating efficient gradient descent during model fine-tuning. We also provide a sparse regularized equivalent form for GPS, which indicates better generalization than full fine-tuning; \romannumeral2) Each neuron within the network possesses the potential to adjust its activation state by fine-tuning selected input connections. Consequently, the pre-trained model exhibits flexibility in modifying features of varying granularities to suit diverse downstream tasks. For instance, when adapting a model pre-trained on ImageNet~\cite{deng2009imagenet} for CIFAR-100~\cite{rebuffi2017learning}, it is necessary to refine high-level features; whereas for ImageNet-Sketch~\cite{wang2019learning} adaptation, more detailed feature fine-tuning is required. \romannumeral3) Our approach avoids introducing extra parameters and computational costs and keeps the architecture of the model intact; \romannumeral4) The selection procedure enables its application across diverse models by adopting a neuron-based rather than a layer-based method, thereby eliminating the necessity for distinct designs for different layers in various models. \romannumeral5) Different from other methods using a pre-defined and consistent strategy for different tasks, our method adaptively selects parameters for each task by our proposed gradient strategy to better fit the domain-specific semantics of different downstream tasks. Please see the difference between our method with others in \cref{fig: differences with other methods} and \cref{table: Comparison between different fine-tuning methods}.

We evaluate our approach on a total of 27 visual tasks (including image classification and semantic segmentation) over 4 different model architectures. Our GPS achieves state-of-the-art performance compared to other PEFT methods and has a good balance between performance and the number of trainable parameters, as illustrated in \cref{fig:comparison}. 
Compared with the full fine-tuning, GPS achieves 3.33\% (FGVC) and 9.61\% (VTAB) improvement of the accuracy while tuning only 0.36\% parameters of the pre-trained model on average over 24 tasks; it also demonstrates
a significant improvement of 17\% and 16.8\% in mDice and
mIoU, respectively, on medical image segmentation task. Moreover, we verify the effectiveness of our approach on different network architectures, such as Transformer and Convolutional Neural Networks. Furthermore, we compare GPS with various parameter selection methods and demonstrate its superior properties. GPS provides a new paradigm for PEFT and inspires deeper insights into this field.
\begin{figure*}[!t]
    \centering
    \includegraphics[width=1\linewidth]{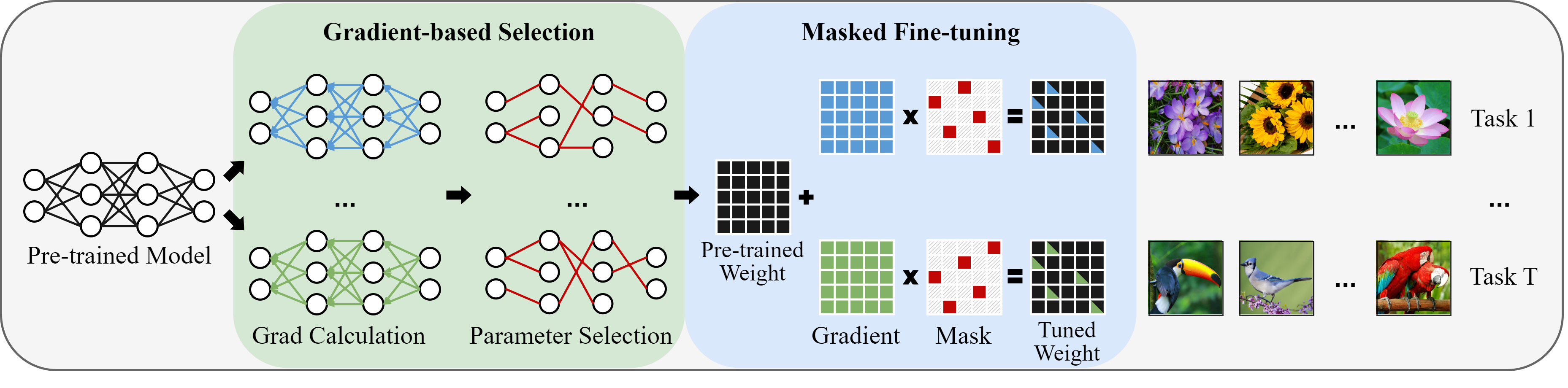}
    \caption{The overall pipeline of GPS. We first select a small portion of important parameters (sub-network) for each task from the original pre-trained model using a gradient-based method. Then only fine-tune the sub-network while keeping other parameters frozen. 
    }
    \label{fig:framework}
\end{figure*}

\section{Related work}
\label{sec:Related Work}

\paragraph{Visual parameter efficient fine-tuning} In general, there are typically two primary categories of PEFT. Addition-based methods introduce additional parameters to the pre-trained backbone. Adapters \cite{houlsby2019parameter,rebuffi2017learning,pfeiffer2020adapterhub,stickland2019bert,pfeiffer2020adapterfusion,sung2022vl,zhang2021tip,wang2020k,zhang2022pointclip,gao2021clip,zhang2023llama,gao2023llama} adopt a residual pathway and learn down and up projection with a nonlinear activation. Others~\cite{mahabadi2021parameter} propose a hyper-network to generate model weights or decompose the dense weighted matrix into the low-rank matrix~\cite{karimi2021compacter}. Prompt methods \cite{ju2022prompting, gao2020making, hu2021knowledgeable, liu2023pre, ding2021openprompt,li2021prefix,liu2022p,zhang2023prompt,zhu2023not} wrap the input with context. VPT~\cite{jia2022visual} prepend learnable prompts to the input tokens. SSF~\cite{lian2022scaling} achieves promising results by scaling and shifting the feature between layers. 
Selection-based methods select a subset of the parameters for tuning, such as only fine-tuning bias~\cite{zaken2021bitfit}, last K layers~\cite{houlsby2019parameter, jia2022visual}. While traditionally considered less effective than addition-based methods, our approach of adaptively selecting parameters for each task yielded surprisingly strong results.

\paragraph{Sub-network training}
Pruning technique \cite{han2015learning,gale2019state,han2015deep,kruschke1991benefits,li2016pruning,wen2016learning} uncovers the importance of subnetworks. The lottery ticket hypothesis~\cite{frankle2018lottery} articulates that subnetworks can reach the accuracy of the original model. Fine-tuning sub-networks are widely studied. SpotTune~\cite{guo2019spottune} designs a policy network to make routing decisions for subset networks. Child-tuning~\cite{xu2021raise} iteratively updates a subset of parameters by masking out some gradients during the backward process. However, these methods are not aligned with the PEFT setting. We fix a small number of parameters and only tune them for fitting downstream to achieve PEFT.


\section{Approach}
\label{approach}
Different from the currently popular methods introducing additional parameters to fine-tune the pre-trained model for downstream tasks \cite{jia2022visual, houlsby2019parameter, lian2022scaling}, we select only a small number of parameters from the pre-trained model and then only update these parameters during the fine-tuning stage. Specifically, our method has two stages: parameter selection and masked fine-tuning. For each downstream task $t$, we first select a small portion of important parameters (task-specific parameters) from the original pre-trained model using a gradient-based method. We then fine-tune the pre-trained model for the task $t$, keeping all other unimportant parameters frozen and updating only selected parameters using a sparse binary mask to set the gradient of unimportant parameters to zero (see \cref{fig:framework}).


\subsection{Gradient-based parameter selection}
Relevant studies have indicated that the pre-trained backbone exhibits diverse feature patterns at distinct parameter positions, and the same positions make varying contributions to fine-tuning various tasks \cite{chatterji2019intriguing, naseer2021intriguing, raghu2021vision, kumar2022fine, yosinski2014transferable}.
Therefore, we posit that there exists an optimal subset of parameters for fine-tuning a pre-trained model to a downstream task. This subset is essential and necessary for fine-tuning the task, and the different tasks require a distinct subset. Formally, given a downstream task $t$ with the dataset $\mathcal{D}_{t}$ and a pre-trained model $\Theta=\left\{w_{1}, w_{2}, \ldots, w_{N}\right\}$, we aim to find a subset of $\boldsymbol{w}$, i.e. $\boldsymbol{w}=$ $\left\{w_{1}, w_{2}, \ldots, w_{n}\right\}$ ($n \ll N$). we select parameters following two principles: 1) Important for downstream tasks; 2) Distributed over the whole network.


\paragraph{Importance for downstream tasks} We identify the importance of parameters in a pre-trained model for a specific task by selecting those with the highest gradient value, which is obtained by calculating the gradient of a loss function with respect to its parameters. The intuition behind this is that the parameters with the largest gradient value indicate the loss function changes fastest along the gradient direction and has the greatest change rate, which facilitates efficient gradient descent during fine-tuning. Specifically, the gradient of the parameters is calculated by
\begin{equation}
\nabla \mathcal{L}_{\mathcal{D}_{t}}(\Theta)=\left[
\frac{\partial \mathcal{L}}{\partial w_1} \\
\cdots \\
\frac{\partial \mathcal{L}}{\partial w_N}
\right]^{\top}
\end{equation}
where $\mathcal{L}(\boldsymbol{w})$ is the loss function. Normally, when we fine-tune a pre-trained model on a downstream task, we need a new classification head ($i.e.$ MLP) with random initialization. In order to avoid the adverse effects of these randomly initialized parameters on gradient calculation using the cross-entropy loss function, we use Supervised Contrastive Loss (SCL)~\cite{Khosla2020SupervisedCL} as the loss function for calculating the gradient during parameter selection, since it does not need to involve the head (We still use cross-entropy loss during fine-tuning stage). SCL is a variant of Contrastive Loss (CL) that aims to bring different augmented samples of the same image closer together in embedding space. In contrast, SCL tries to cluster samples from the same class together, which coincides with our target of the downstream classification tasks. Specifically, given a task with the dataset $\mathcal{D}_{t} = $ $\left\{\boldsymbol{x}_{i}, y_{i}\right\}_{i=1 \ldots K}$, SCL is calculated by
\begin{equation}
\begin{split}
&\mathcal{L}^{\text {scl}}=\sum_{i \in \mathcal{D}_{t}} \mathcal{L}_{i}^{\text {scl}}\\
&=\sum_{i \in \mathcal{D}_{t}} \frac{-1}{|P(i)|} \sum_{p \in P(i)} \log \frac{\exp \left(\boldsymbol{z}_{i} \odot \boldsymbol{z}_{p} / \tau\right)}{\sum_{a \in A(i)} \exp \left(\boldsymbol{z}_{i} \odot \boldsymbol{z}_{a} / \tau\right)}
\end{split}
\end{equation}
where $i$ represents $i^{th}$ sample in $\mathcal{D}_{t}$; $P(i) \equiv\left\{p \in A(i): \tilde{y}_{p}=\tilde{y}_{i}\right\}$ is subset of $\mathcal{D}_{t}$, in which all samples have the same class with $i$; $A(i) \equiv \mathcal{D}_{t}  \backslash\{i\}$; $\boldsymbol{z}$ is the feature extacted from the pre-trained encoder and $\tau \in \mathcal{R}^{+}$ is a scalar temperature parameter. 

\paragraph{Equivalent with sparse regularization} In the above, we implicitly assume that the order of $\left(\left\|\frac{\partial \mathcal{L}}{\partial w_1}\right\| \cdots \left\|\frac{\partial \mathcal{L}}{\partial w_N}\right\| \right)$ is the same as $\left( \left\|w_1' - w_1\right\| \cdots \left\|w_N' - w_N\right\| \right)$, which means selecting parameters with top-n gradient norm is the same as selecting top-n of the fine-tuning changes. Therefore, GPS captures the top-n important parameters for downstream tasks. The optimization objective can be rewritten as
\begin{equation}
\Theta' = \min \mathcal{L}(\Theta') \qquad   
   s.t. \quad \|\Theta' - \Theta\|_0 \leq n
\end{equation}
where $\|\|_0$ is the $l_0$ norm and $\Theta'$ is the fine-tuned model. By Lagrangian duality, solving the above problem is equivalent to solving the following problem:
\begin{equation}\label{eq:spare regularization}
    \qquad \mathcal{L}(\Theta') + \lambda\|\Theta' - \Theta\|_0
\end{equation}
with appropriate $\lambda$. Hence, GPS can be reviewed as a sparse regularized fine-tuning, which may lead to better generalization. \citet{fu2023effectiveness} demonstrate that \cref{eq:spare regularization} has smaller generalization bound than pure optimization toward $\mathcal{L}$ with full fine-tuning, resulting in better performance.

\paragraph{Distribution over the whole network}
A simple idea for parameter selection is to select a certain percentage of parameters with the highest gradient from the entire network. Our experiments have shown that with this idea, the majority of the selected parameters are located in the top layers of the network (see Supplementary for details), which is consistent with the findings reported in \cite{houlsby2019parameter, Howard2018UniversalLM}. However, solely fine-tuning these top-layer parameters is insufficient to mitigate the impact of the pre-trained model's own inductive bias, particularly when there exist substantial disparities in data distributions between upstream and downstream tasks, which need to fine-tune more detailed features from shallower layers. Motivated by various studies indicating the distinct roles played by different components of neural networks \cite{Wang2021PACBayesIB, Proakis1992DigitalSP, Shrikumar2017LearningIF, Cao2022TowardsID, Fan2020OnIO}, we posit that when fine-tuning a pre-trained model for downstream tasks, the adjusted parameters should be distributed throughout the entire network. The reason behind this lies in the ability of the model to adapt the information stored in parameters at different levels of granularity to fit downstream tasks. Therefore, our strategy is that for each neuron in the network, we select top-K (at least one) connections (weights) among all the input connections of the neuron, as shown in \cref{fig:framework}. By doing so, every neuron within the network possesses the potential to fine-tune its activation state rather than solely adjust high-level information in the top layers. In other words, our approach fine-tunes the detailed information stored in each neuron of the network, which better fits the downstream task during the fine-tuning stage. Our exploratory experiment further substantiates this assertion, as shown in \cref{tab:ablation}(a). 

Combining the two points above,  we first calculate the gradient of the loss with respect to all the weights in the models for a specific task. Then for each neuron in the network, we select top-K connections with the highest gradient value (the modulus of gradient) among all input connections to the neuron. Doing so can not only ensure that important parameters for downstream tasks are chosen and allow the model to tune the activation state of all neurons for better fitting of downstream tasks. Another benefit of this selection procedure is its ease of application across various model architectures, such as Transformer and CNN, avoiding any model-specific design. Our experiments also demonstrate the effectiveness of our approach across diverse architectures, as shown in \cref{tab:fgvc} and \cref{tab:architecture}. 

\subsection{Masked fine-tuning} 
After parameter selection for a specific task, we fine-tune the pre-trained model on the task. During fine-tuning, we only update the selected parameters while keeping the remaining parameters of the pre-trained model frozen.  As our selected parameters are distributed across all neurons in every layer, only a few parameters within a specific weight matrix of the network are chosen, resulting in the updated matrix being sparse. Therefore, we utilize a mask to help with the sparse training. Specifically, for $j^{th}$ weight matrix $\boldsymbol{W}_j \in \mathbb{R}^{d_{\text {in }} \times d_{\text {out }}}$ in the network, we build a same size of binary mask $\boldsymbol{M}_j \in \mathbb{R}^{d_{\text {in }} \times d_{\text {out }}}$:
\begin{equation}
\boldsymbol{M}_{j}=\left\{\begin{array}{ll}
1, & w_{j}^{k} \in \boldsymbol{w} \\
0, & w_{j}^{k} \notin \boldsymbol{w}
\end{array}\right.
\end{equation}
where $w_{j}^{k}$ represents $k^{th}$ element in $j^{th}$ weight matrix. For each element in $\boldsymbol{M}_{j}$, its value is set to 1 if the corresponding parameter in $\boldsymbol{W}_j$ is selected, and 0 otherwise. Then the weight matrix is updated by
\begin{equation}
\boldsymbol{W}_j \leftarrow \boldsymbol{W}_j-\epsilon 
\nabla \mathcal{L}(\boldsymbol{W}_j) \odot \boldsymbol{M}_j
\end{equation}
where $\nabla \mathcal{L}(\boldsymbol{W}_j)$ 
is the gradient of the cross-entropy loss with respect to $\boldsymbol{W}_j$. As a result, the gradients of unselected parameters are zeroed out and excluded from updates, while only a small number of our selected parameters are updated during fine-tuning for downstream tasks. Please see \cref{fig:framework} for a visualization of our method.
\section{Experiments}
\label{experiments}

We evaluate GPS on various downstream tasks, including image classification tasks and semantic segmentation tasks with different architectures. First, we briefly introduce our experimental settings, including datasets, backbones, and baselines. Then we demonstrate the effectiveness and universality of GPS. Moreover, we systematically study the impacts of different selection schemes and conduct comprehensive ablation experiments.

\subsection{Experimental settings}
\paragraph{Datasets} Following VPT~\cite{jia2022visual} and SSF~\cite{lian2022scaling}, we evaluate our GPS method on a series of datasets categorized into three groups: 
\romannumeral1) \textbf{\textit{FGVC}}: Fine-Grained Visual Classification (FGVC) benchmark includes 5 downstream tasks, which are CUB-200-2011~\cite{wahcaltech}, NABirds~\cite{van2015building}, Oxford Flowers~\cite{nilsback2008automated}, Stanford Dogs~\cite{khosla2011novel} and Stanford Cars~\cite{gebru2017fine}. 
\romannumeral2) \textbf{\textit{VTAB-1k}}: Visual Task Adaptation Benchmark~\cite{zhai2019large} (VTAB) contains 19 visual classification tasks which are grouped into three sets: Natural, Specialized, and Structured.
\romannumeral3) \textbf{\textit{CIFAR-100}}~\cite{krizhevsky2009learning} and \textbf{\textit{ImageNet-1k}}~\cite{deng2009imagenet}: widely use for general image classification task. 

\begin{table}[!t]
    \centering
    \resizebox{1\linewidth}{!}{
        \begin{tabular}{c|ccccccc}
        \toprule
        Dataset & \makecell[c]{CUB\\-2011} & \makecell[c]{NA-\\Brids} & \makecell[c]{Oxford\\Flowers} & \makecell[c]{Stan.\\Dogs} & \makecell[c]{Stan.\\Cars} & \makecell[c]{Mean\\Acc.} & \makecell[c]{Params.\\(\%)} \\
        \midrule
        Full~\cite{jia2022visual}         & 87.3         & 82.7    & 98.8           & 89.4          & 84.5          & 88.54     & 100.00           \\
        \midrule
        Linear~\cite{jia2022visual}         & 85.3         & 75.9    & 97.9           & 86.2          & 51.3          & 79.32     & 0.21             \\
        Bias~\cite{zaken2021bitfit}           & 88.4         & 84.2    & 98.8           & 91.2          & 79.4          & 88.40     & 0.33             \\
        \midrule
        Adapter~\cite{houlsby2019parameter}       & 87.1         & 84.3    & 98.5           & 89.8          & 68.6          & 85.66     & 0.48             \\
        LoRA~\cite{hu2021lora}           & 85.6         & 79.8    & 98.9           & 87.6          & 72.0          & 84.78     & 0.90             \\
        VPT-Shallow~\cite{jia2022visual}    & 86.7         & 78.8    & 98.4           & 90.7          & 68.7          & 84.62     & 0.29             \\
        VPT-Deep~\cite{jia2022visual}       & 88.5         & 84.2    & 99.0           & 90.2          & 83.6          & 89.11     & 0.99             \\
        SSF~\cite{lian2022scaling}            & \underline{89.5}         & \underline{85.7}    & \underline{99.6}           & 89.6          & \underline{89.2}          & \underline{90.72}     & 0.45             \\
        SPT-Adapter~\cite{he2023sensitivity}    & 89.1         & 83.3    & 99.2           & 91.1          & 86.2          & 89.78     & 0.47             \\
        SPT-LoRA~\cite{he2023sensitivity}       & 88.6         & 83.4    & 99.5           & \underline{91.4}          & 87.3          & 90.04     & 0.60             \\
        \midrule
        GPS (Ours)     & \textbf{89.9}         & \textbf{86.7}    & \textbf{99.7}           & \textbf{92.2}          & \textbf{90.4}          & \textbf{91.78}     & 0.77             \\
        \bottomrule
        \end{tabular}
    }
    \caption{Performance comparisons on FGVC with ViT-B/16 models pre-trained on ImageNet-21K.}
    \label{tab:fgvc}
\end{table}

\begin{table*}[!t]
    \centering
    \resizebox{2.1\columnwidth}{!}{
        \begin{tabular}{c|ccccccc|cccc|cccccccc|cc}
        \toprule
         \multirow{2}{*}{\diagbox[height=8\line]{Method \\ \\ \\}{\\ \\ \\ Dataset}} & \multicolumn{7}{c}{Natural}                                       & \multicolumn{4}{c}{Specialized}                   & \multicolumn{8}{c}{Structured}                                                                                      & \multicolumn{2}{c}{VTAB}     \\
        \cline{2-22}
         & \rotatebox{90}{CIFAR-100} & \rotatebox{90}{Caltech101} & \rotatebox{90}{DTD} & \rotatebox{90}{Flowers102} & \rotatebox{90}{Pets} & \rotatebox{90}{SVHN} & \rotatebox{90}{Sun397} & \rotatebox{90}{Patch Camelyon} & \rotatebox{90}{EuroSAT} & \rotatebox{90}{Resisc45} & \rotatebox{90}{Retinopathy} & \rotatebox{90}{Clevr/count} & \rotatebox{90}{Clevr/distance} & \rotatebox{90}{DMLab} & \rotatebox{90}{KITTI/distance} & \rotatebox{90}{dSprites/loc} & \rotatebox{90}{dSprites/ori} & \rotatebox{90}{SmallNORB/azi} & \rotatebox{90}{SmallNORB/ele} & \rotatebox{90}{Mean Acc.} & \rotatebox{90}{Mean Params. (\%)} \\
        \midrule
        Full~\cite{jia2022visual}                            & 68.9      & 87.7       & 64.3 & 97.2       & 86.9 & 87.4 & 38.8   & 79.7           & 95.7    & 84.2     & 73.9        & 56.3        & 58.6           & 41.7  & 65.5           & 57.5         & 46.7         & 25.7          & 29.1          & 65.57     & 100.00           \\
        \midrule
        Linear~\cite{jia2022visual}                          & 63.4      & 85.0       & 64.3 & 97.0       & 86.3 & 36.6 & 51.0   & 78.5           & 87.5    & 68.6     & 74.0        & 34.3        & 30.6           & 33.2  & 55.4           & 12.5         & 20.0         & 9.6           & 19.2          & 53.00     & 0.05             \\
        Bias~\cite{zaken2021bitfit}                            & 72.8      & 87.0       & 59.2 & 97.5       & 85.3 & 59.9 & 51.4   & 78.7           & 91.6    & 72.9     & 69.8        & 61.5        & 55.6           & 32.4  & 55.9           & 66.6         & 40.0         & 15.7          & 25.1          & 62.05     & 0.16             \\
        \midrule
        Adapter~\cite{houlsby2019parameter}                         & 74.1      & 86.1       & 63.2 & 97.7       & 87.0 & 34.6 & 50.8   & 76.3           & 88.0    & 73.1     & 70.5        & 45.7        & 37.4           & 31.2  & 53.2           & 30.3         & 25.4         & 13.8          & 22.1          & 55.82     & 0.31             \\
        LoRA~\cite{hu2021lora}                         & 68.1      & 91.4       & 69.8 & 99.0       & 90.5 & 86.4 & 53.1   & 85.1           & 95.8    & 84.7     & 74.2        & \underline{83.0}        & 66.9           & 50.4  & 81.4           & 80.2         & 46.6         & 32.2          & 41.1          & 72.63     & 0.90             \\
        VPT-Shallow~\cite{jia2022visual}                     & 77.7      & 86.9       & 62.6 & 97.5       & 87.3 & 74.5 & 51.2   & 78.2           & 92.0    & 75.6     & 72.9        & 50.5        & 58.6           & 40.5  & 67.1           & 68.7         & 36.1         & 20.2          & 34.1          & 64.85     & 0.13             \\
        VPT-Deep~\cite{jia2022visual}                        & \underline{78.8}      & 90.8       & 65.8 & 98.0       & 88.3 & 78.1 & 49.6   & 81.8           & \underline{96.1}    & 83.4     & 68.4        & 68.5        & 60.0           & 46.5  & 72.8           & 73.6         & 47.9         & \underline{32.9}          & 37.8          & 69.43     & 0.70             \\
        SSF~\cite{lian2022scaling}                             & 69.0      & 92.6       & \underline{75.1} & \textbf{99.4}       & \textbf{91.8} & \underline{90.2} & 52.9   & \underline{87.4}           & 95.9    & \textbf{87.4}     & 75.5        & 75.9        & 62.3           & \underline{53.3}  & 80.6           & 77.3         & \underline{54.9}         & 29.5          & 37.9          & 73.10     & 0.28             \\
        SPT-ADAPTER~\cite{he2023sensitivity}                     & 72.9      & 93.2       & 72.5 & \underline{99.3}       & 91.4 & 88.8 & \textbf{55.8}   & 86.2           & \underline{96.1}    & 85.5     & 75.5        & \underline{83.0}        & \textbf{68.0}           & 51.9  & 81.2           & 51.9         & 31.7         & \textbf{41.2}          & \textbf{61.4}          & 73.03     & 0.44             \\
        SPT-LoRA~\cite{he2023sensitivity}                        & 73.5      & \underline{93.3}       & 72.5 & \underline{99.3}       & 91.5 & 87.9 & \underline{55.5}   & 85.7           & \textbf{96.2}    & 85.9     & \underline{75.9}        & \textbf{84.4}        & \underline{67.6}           & 52.5  & \underline{82.0}           & \underline{81.0}         & 51.1         & 30.2          & 41.3          & \underline{74.07}     & 0.63             \\
        \midrule
        GPS (Ours)                        & \textbf{81.1}      & \textbf{94.2}       & \textbf{75.8} & \textbf{99.4}       & \underline{91.7} & \textbf{91.6} & 52.4   & \textbf{87.9}           & \textbf{96.2}    & \underline{86.5}     & \textbf{76.5}        & 79.9        & 62.6           & \textbf{55.0}  & \textbf{82.4}           & \textbf{84.0}         & \textbf{55.4}         & 29.7          & \underline{46.1}          & \textbf{75.18}     & 0.25              \\
        \bottomrule
        \end{tabular}
    }
    \caption{Performance comparisons on VTAB-1k with ViT-B/16 models pre-trained on ImageNet-21K.}
    \label{tab:vtab}
\end{table*}

\paragraph{Backbones} For a fair comparison, we follow VPT and SSF by using ViT-B/16~\cite{dosovitskiy2020image} pre-trained on ImageNet-21K~\cite{deng2009imagenet} for the main image classification experiments. Moreover, to demonstrate the universality of our GPS, we also explore other backbones, including Swin Transformer~\cite{liu2021swin} and ConvNeXt-B~\cite{liu2022convnet} for another variant of Transformer-based and CNN-based architecture, respectively. In addition, we finetune
semantic segmentation tasks on SAM~\cite{kirillov2023segany}, a strong segmentation foundation model.

\paragraph{Baselines} We compare our GPS with a variety of fine-tuning protocols that can be mainly categorized into three types:
\romannumeral1) \textbf{\textit{Full}}: Full fine-tuning is the most commonly used protocol updating all parameters of the whole model during tuning.
\romannumeral2) \textbf{\textit{Selection-based}}: This kind of method selects a subset of parameters in the original model for fine-tuning, including linear probing and Bias~\cite{zaken2021bitfit}. Such methods are easy to implement and require no extra computations but have not worked well. Our method belongs to this group and achieves the best performance while ensuring convenience and universality.
\romannumeral3) \textbf{\textit{Addition-based}}: This kind of method adds new trainable parameters to the backbone, including Adapter~\cite{houlsby2019parameter}, VPT~\cite{jia2022visual} and SPT-Adapter~\cite{he2023sensitivity}. Such methods require extra computations in both the training and inference stages. Other methods like LoRA~\cite{hu2021lora}, SSF~\cite{lian2022scaling}, and SPT-LoRA~\cite{he2023sensitivity} also add new tunable parameters during the training stage, but these parameters can be reparameterized into the backbone during testing.  

\paragraph{Implementation details} We follow SSF to process the images in all the FGVC, VTAB-1k and CIFAR-100 datasets. We employ the Adam~\cite{kingma2014adam} optimizer with cosine learning rate decay to fine-tune models for 100 epochs, and the linear warm-up is used in the first 10 epochs.  All experiments are conducted on the NVIDIA A100 GPU.

\subsection{Performance on image classification}
We present a comprehensive evaluation of the effectiveness of our GPS by comparing it against multiple baselines on 3 benchmarks, comprising a total of 26 datasets. In addition to common benchmarks (FGVC and VTAB-1k), we also compare our method with others on different architectures. We evaluate the performance and effectiveness by Top-1 accuracy (\%) and the number of fine-tuned parameters.

\paragraph{Image classification performance} As shown in \cref{tab:fgvc} and \cref{tab:vtab}, our GPS outperforms all other fine-tuning methods by a large margin on both FGVC and VTAB benchmarks, sufficiently demonstrating that our method of parameter selection is a simple yet effective way for model tuning. On FGVC, GPS outperforms all other fine-tuning methods, including full fine-tuning, on all 5 tasks. It obtains 1.02\% and 3.24\% accuracy improvement of the mean accuracy compared to the previous SOAT method SSF~\cite{lian2022scaling} and full fine-tuning, while it only uses 0.77\% of trainable parameters. On VTAB, GPS also outperforms all other fine-tuning methods. Specifically, it obtains 1.11\% and 9.61\% improvement of the mean accuracy on 19 VTAB tasks compared to the previous SOAT method SPT-LoRA~\cite{he2023sensitivity} and full fine-tuning. GPS beats the previous SOTA by 1.75\%, 0.23\%, and 0.63\% in the Natural, Specialized and Structured subsets, respectively. Meanwhile, GPS also uses fewer trainable parameters compared to VPT-Deep, SSF, and SPT-LoRA (0.25\% vs. 0.70\%, 0.28\% and 0.63\%), which further illustrates the high efficiency of our approach. For most tasks, we exclusively select the top 1 input connection for each neuron; however, for more challenging tasks, multiple connections are chosen (see Supplement for details). The percentage of learnable parameters in our GPS can be explicitly controlled by adjusting the number of connections selected, allowing for a balance between parameter count and performance in tasks.

\begin{table}[!t]
    \centering
    \resizebox{0.9\linewidth}{!}{
        \begin{tabular}{c|cc|cc}
        \toprule
        \multirow{2}{*}{Architecture} & \multicolumn{2}{c}{Swin-B} & \multicolumn{2}{c}{ConvNeXt-B} \\
        \cmidrule{2-5}
                                    & Ave. Acc.      & Params.(\%)     & Ave. Acc.        & Params.(\%) \\
        \midrule
        Full~\cite{jia2022visual}                        & 92.42     & 100.00         & 93.04       & 100.00           \\
        Linear~\cite{jia2022visual}                      & 87.90     & 0.28           & 88.00       & 0.28             \\
        SSF~\cite{lian2022scaling}                         & 91.54     & 0.56           & 92.48       & 0.56             \\
        GPS (Ours)                  & \textbf{92.56}    & 0.95           & \textbf{93.32}      & 0.90             \\
        \bottomrule
        \end{tabular}
    }
    \caption{Performance comparisons on FGVC benchmark (Average accuracy over 5 tasks) with different model architectures.}
    \label{tab:architecture}
\end{table}
\paragraph{Generalization on different architectures} 
Since our method only selects a subset of parameters from the pre-trained model for fine-tuning, it is naturally model-agnostic. We compare GPS with other representative methods across ViT-B/16 (\cref{tab:fgvc}), Swin-B and ConvNeXt-B architectures on the FGVC dataset (\cref{tab:architecture}), CIFAR-100 and ImageNet-1k (Please see full results in Supplementary). 
Among all three architectures, GPS consistently outperforms all other baselines, demonstrating its model-agnostic advantage. The Swin and Convnext have more complex designs than ViT, enabling them to acquire comprehensive and high-quality features during pre-training. Consequently, even the simplest linear probing method yields commendable results on these two architectures, reducing the effectiveness of the PEFT method and causing the previous SOTA SSF to underperform Full. However, in this scenario, our GPS still maintains a lead over Full with gains of 0.12\% and 0.28\%, respectively, further showing the effectiveness of our method.


\begin{figure*}[tpbh]
    \centering
    \subfigure{\includegraphics[width=0.24\linewidth]{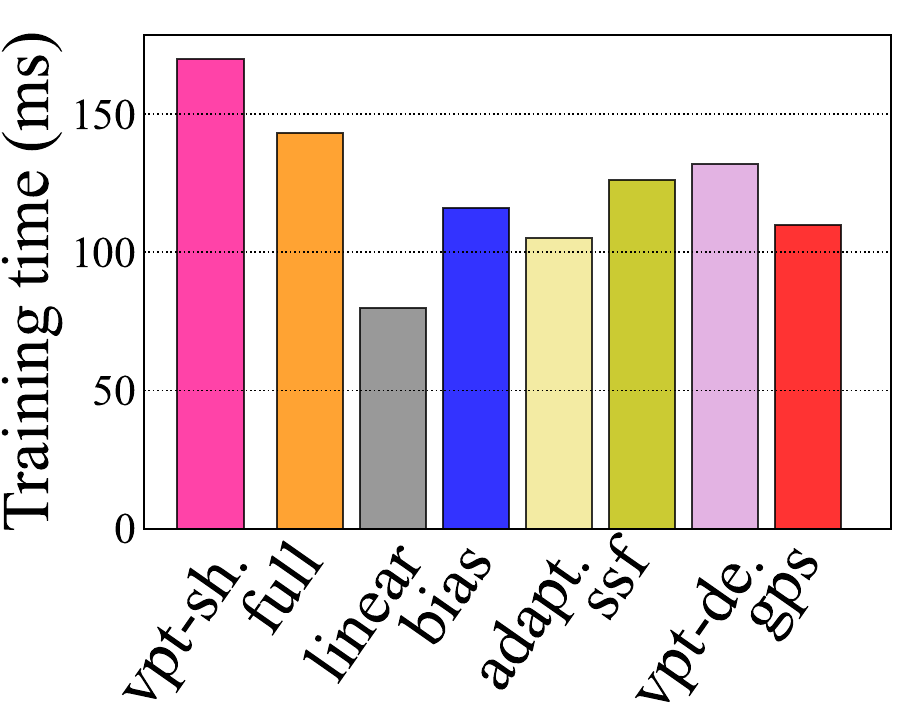}}
    \subfigure{\includegraphics[width=0.24\linewidth]{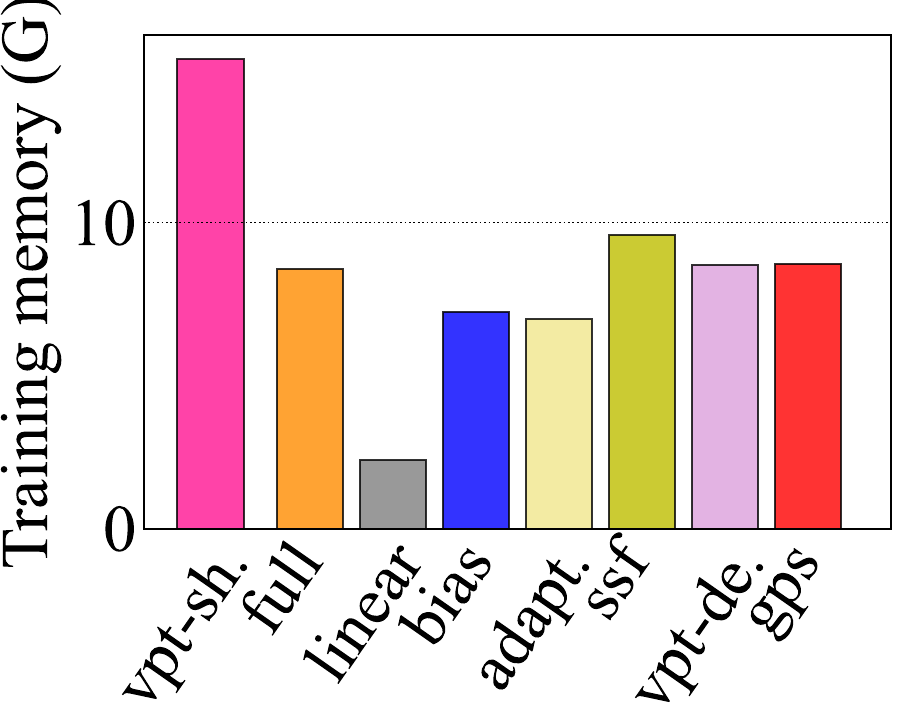}}
    \subfigure{\includegraphics[width=0.24\linewidth]{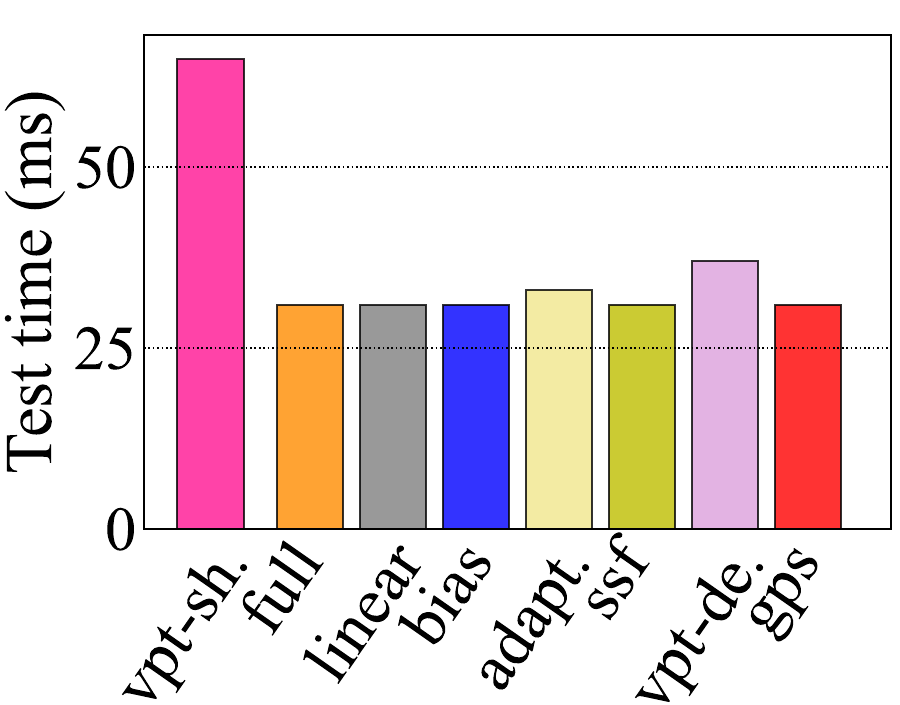}}
    \subfigure{\includegraphics[width=0.24\linewidth]{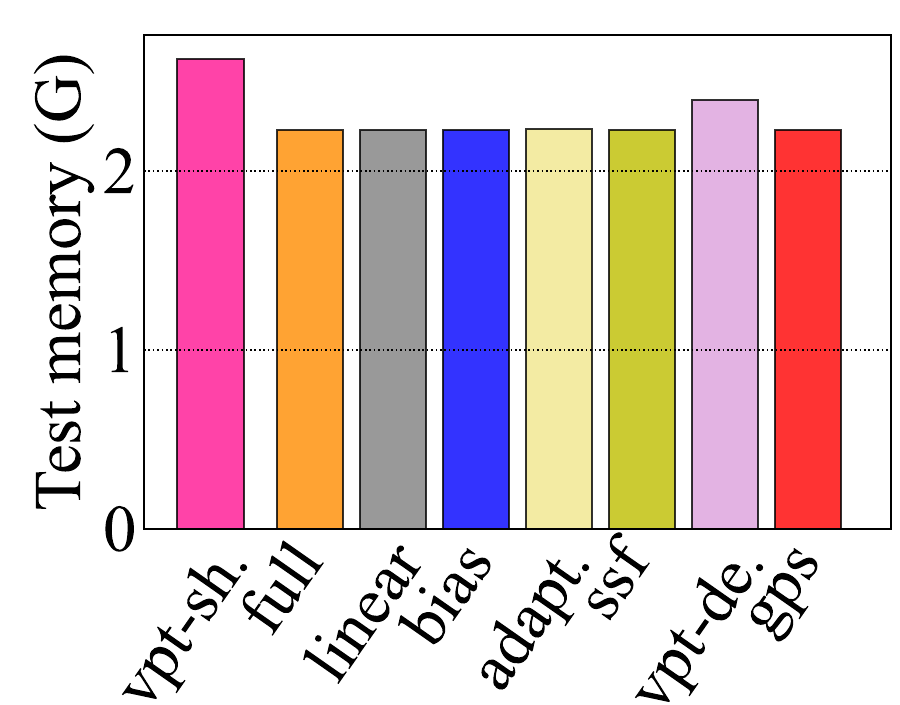}}
     \caption{ Computational cost of different tuning methods. From left to right: training time, training memory, test time, and test memory. Training/Test time is the time consumed by a mini-batch.}
     \label{fig:cost}
\end{figure*}

\paragraph{Computational cost} In \cref{fig:cost}, we compare the computational cost of GPS with other fine-tuning methods to demonstrate the efficiency of our approach. Following SSF~\cite{lian2022scaling}, we reimplement VPT~\cite{jia2022visual} with 200 and 50 prompts for the shallow and
deep versions, respectively. A batch size of 32 is used in both the training and inference stages. For a fair comparison, for all experiments, we do not use mixed precision training, which was used in SSF. All metrics are measured on a single NVIDIA A100 GPU. In the training stage, GPS has less time and memory consumption than both VPT and SSF. Compared with full fine-tuning, GPS has a much lower time overhead and a similar memory overhead, but it leads to an increased performance by a large margin. Since GPS is a selection-based method, it does not introduce any additional parameters, so it can achieve the same minimum time and memory overhead as full fine-tuning during inference without any reparameterization operation, which is much lower than the addition-based Adapter and VPT.
\begin{table}
	\centering
 \resizebox{0.8\linewidth}{!}{
	\begin{tabular}{c|cc|c}
		\toprule
		Method & mDice ($\uparrow$) & mIoU ($\uparrow$) &Params. (M)  \\ 
		\midrule
	Full~\cite{jia2022visual} & 71.1 & 55.7 & 93.8  \\ 
        \midrule
        Linear~\cite{jia2022visual} & 71.6    & 46.6  &4.06  \\
        Bias~\cite{zaken2021bitfit} & 86.5 & 69.1 & 4.16 \\
        \midrule
	Adapter~\cite{chen2023sam} & 84.8 & 66.7 & 4.12\\
        SSF~\cite{lian2022scaling} & 87.3 & 71.7 & 4.26\\ 
        \midrule
        GPS (Ours)  & \textbf{88.1}  & \textbf{72.5} & 4.22 \\ \bottomrule
	\end{tabular}}
	 \caption{Quantitative Result for Polyp Segmentation}
    \label{tab:polyp-base}
\end{table}
\subsection{Semantic segmentation}
\label{Semantic Segmentation}
In addition to visual classification tasks, we also explore our method for the task of semantic segmentation. Segment Anything Model (SAM)~\cite{kirillov2023segany} is a strong foundation model for segmentation. It is pre-trained on a large-scale dataset enabling powerful generalization. However, several studies, e.g.~\cite{chen2023sam}, have reported poor performance of SAM on medical segmentation tasks such as polyp segmentation~\cite{jha2020kvasirseg}. To address this limitation, they proposed employing Adapter to effectively fine-tune SAM for downstream medical segmentation tasks. Following their experimental setup, we applied our method to SAM and conducted a comparative analysis with other PEFT approaches. Our GPS yielded exceptional results, as shown in~\cref{tab:polyp-base} and visually depicted in ~\cref{fig:polyp-base} (See Supplementary for more case visualization).

\begin{figure}[!t]
    \centering
    \includegraphics[width=0.9\linewidth]{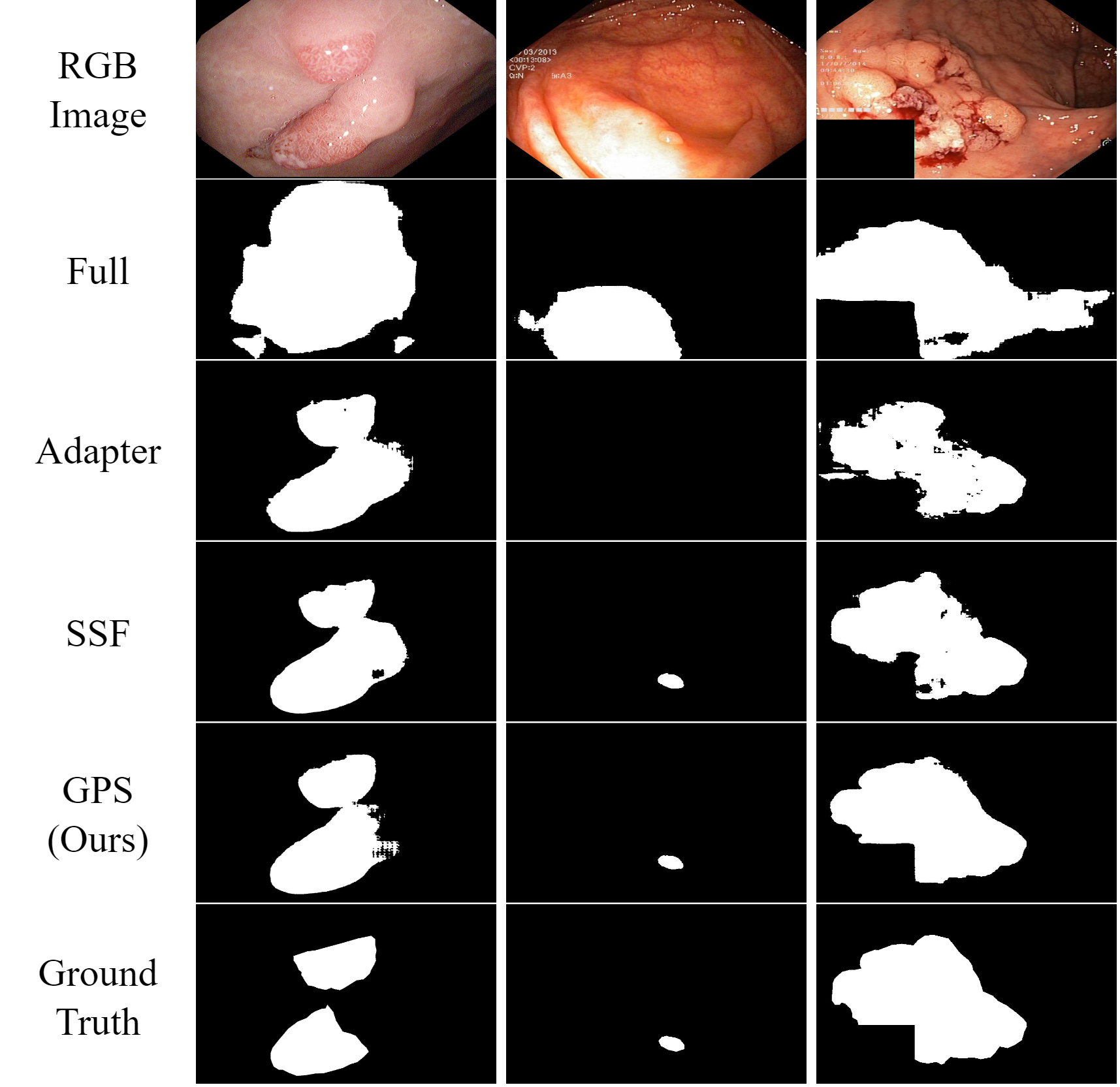}
\caption{The Visualization of Polyp segmentation task. Our GPS can still handle difficult segmentation cases compared with others.}
    \label{fig:polyp-base}
\end{figure}

\subsection{Impacts of different selection schemes}

\begin{figure*}[!t]
    \centering
    \subfigure{\includegraphics[width=0.3\linewidth]{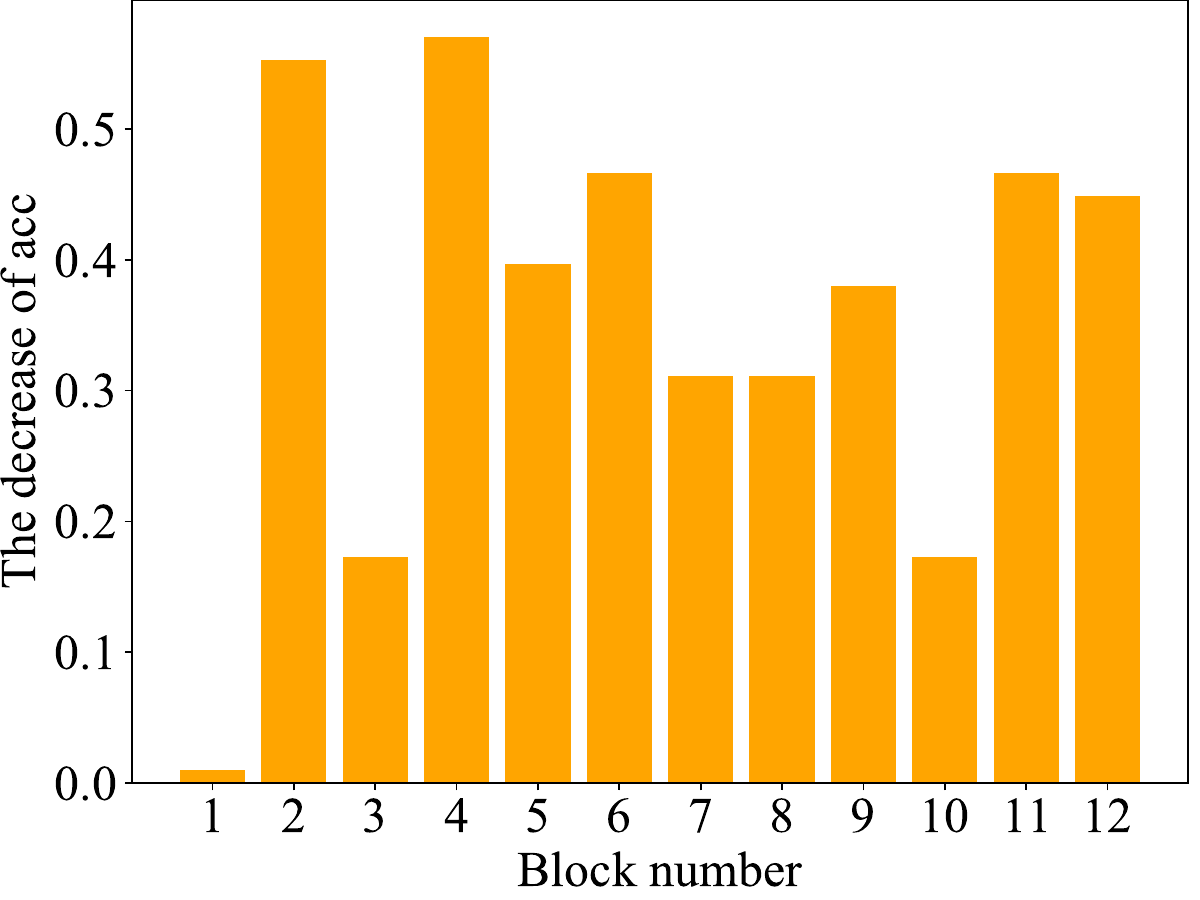}
    \label{fig:ablation:Ablation_block_index}}
    \subfigure{\includegraphics[width=0.3\linewidth]{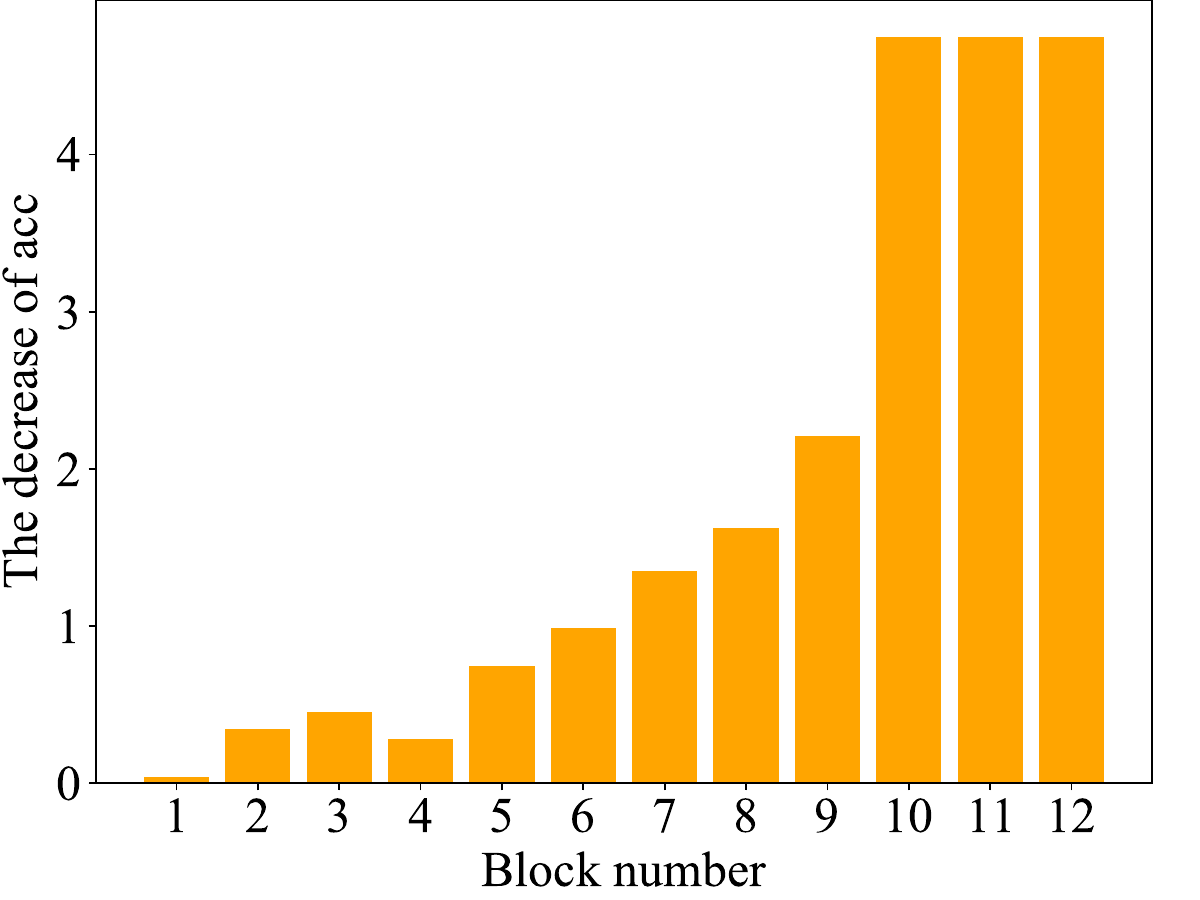}
    \label{fig:ablation:Ablation_block_top}}
    \subfigure{\includegraphics[width=0.31\linewidth]{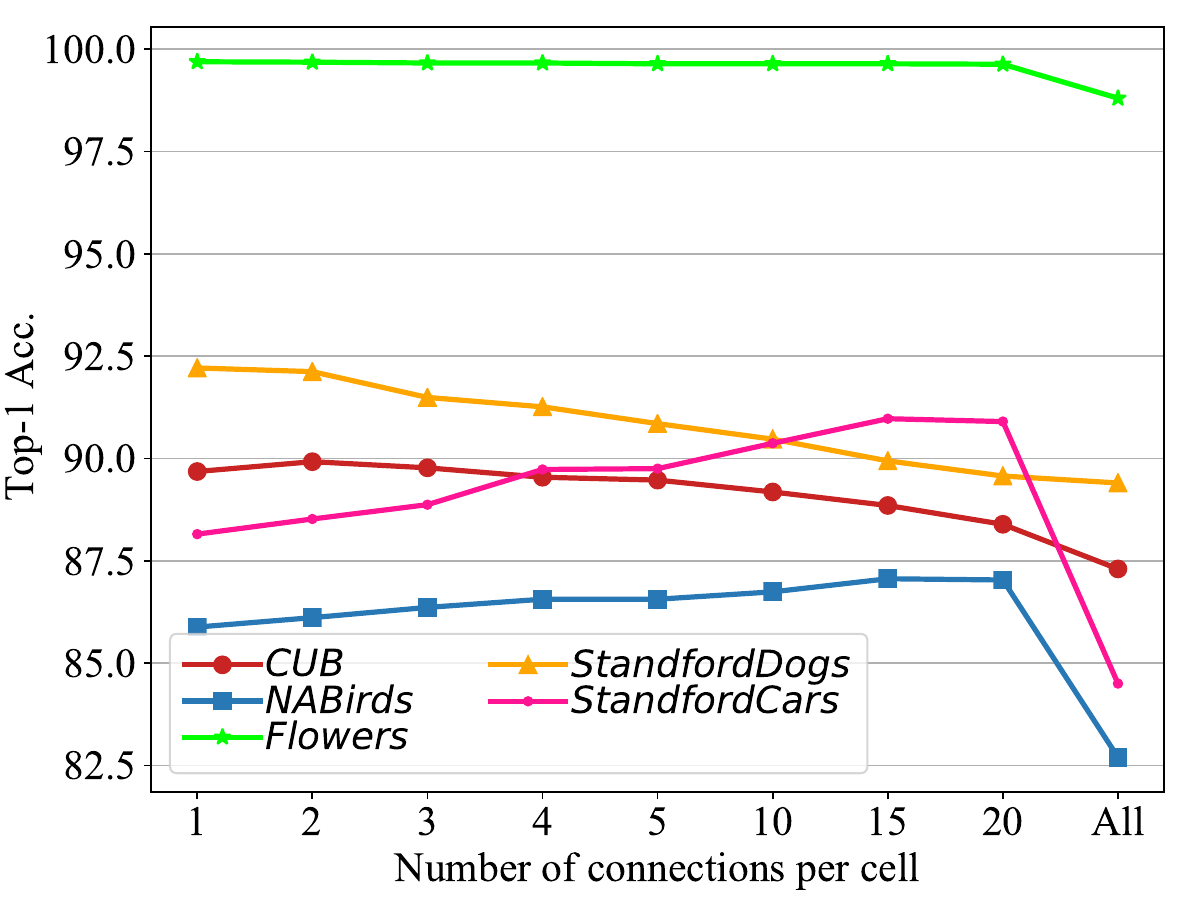}
    \label{fig:ablation:Figure_Ablation_Num}}
    \caption{Impacts of different selection locations and quantities. From left to right: (a) Performance drop caused by not selecting parameters from k-th blocks. (b) And by not selecting from the top k blocks. (c) Impacts of different numbers of selected connections on performance.}
    \label{fig:ablation}
\end{figure*}

\paragraph{Different selection levels} Our GPS selects trainable parameters at the neuron level, i.e. selecting top-k input connection per neuron. We also investigate parameter selection methods at different levels. As shown in \cref{tab:ablation} (a), \textit{Net} and \textit{Layer} represent selecting a certain proportion of the parameters with the highest gradient based on the entire network and each layer, respectively. For a fair comparison, we keep the same number of parameters selected over these levels. We can see that the finer the granularity of selection, the better the performance. For example, the accuracy on CUB increases by 0.44\% and 0.77\% when selection level changes from network to layer, and from layer to neuron.

\paragraph{Different selection criteria} We further study the effectiveness of our gradient-based selection method by comparing different selection criteria. As shown in \cref{tab:ablation} (b), \textit{Net} \textit{Random} and \textit{Neuron Random} means randomly selecting top-K the input connection for each neuron and selecting the same number of parameters based on the whole network respectively. \textit{Magnitude} means selecting top-K input connections with the largest weight per neuron. As we can see, the increase in the randomness of parameter selection causes a decrease in performance (\textit{Net Random}\textless \textit{Neuron Random}). The result of \textit{Magnitude} is similar to \textit{Neuron Random}, demonstrating neuron-level selection is crucial.

\begin{table}[]
\centering
\resizebox{0.9\linewidth}{!}{
\begin{tabular}{ll|lllll}
\toprule
                                   &  Dataset   & CUB   & NAbirds & Flowers & Cars  & Dogs  \\ \midrule
\multicolumn{1}{l|}{\multirow{2}{*}{(a)}} & Net           & 86.86 & 86.55   & 99.62   & 89.65 & 91.32 \\
\multicolumn{1}{l|}{}                     & Layer         & 87.30 & 86.79   & 99.64   & 90.03 & 91.90 \\ \midrule
\multicolumn{1}{l|}{\multirow{3}{*}{(b)}} & Net Random    & 86.60 & 85.98   & 99.61   & 89.10 & 91.34 \\
\multicolumn{1}{l|}{}                     & Neuron Random & 87.17 & 86.02   & 99.62   & 89.52 & 91.82 \\
\multicolumn{1}{l|}{}                     & Magnitude     & 87.29 & 85.99   & 99.62   & 89.29 & 91.30 \\ \midrule
\multicolumn{1}{l|}{(c)}                  & Head+CE       & 87.05 & 86.20   & 99.64   & 89.25 & 91.29 \\ \midrule
&GPS           & \begin{tabular}[c]{@{}l@{}}\textbf{88.07}\\ ±0.11\end{tabular}  & \begin{tabular}[c]{@{}l@{}}\textbf{86.64}\\ ±0.03\end{tabular}  & \begin{tabular}[c]{@{}l@{}}\textbf{99.69}\\ ±0.01\end{tabular}  & \begin{tabular}[c]{@{}l@{}}\textbf{90.10}\\ ±0.10\end{tabular}  & \begin{tabular}[c]{@{}l@{}}\textbf{92.30}\\ ±0.10\end{tabular}  \\ \bottomrule
\end{tabular}}
\caption{The result on FGVC for investigating impacts of different selection schemes and ablations. (a) Different selection levels. (b) Different selection criteria. (c) Gradient calculating method.}
\label{tab:ablation}
\end{table}

\paragraph{Different selection location}
To investigate the impact of selected parameters located at different layers within the network, we conducted experiments using the ViT-B/16 model fine-tuning on CUB and evaluated accuracy degradation when applying our GPS method to select parameters from the entire network except for a specific transformer block or previous several transformer blocks. 
As shown in \cref{fig:ablation:Ablation_block_index}, it is surprising to note that when we do not select parameters from a specific block, the biggest drop in the accuracy comes from the shallow layers (block 2 and block 4). This finding supports our GPS approach that selects parameters from the entire network rather than just the last few layers. When we do not select the parameters from the first specific number of blocks, it is observed that the accuracy drop increases with more blocks removed (\cref{fig:ablation:Ablation_block_top}).

\subsection{Ablation study}

\paragraph{Head-free contrastive loss} 
To obtain more accurate gradients for selecting parameters, inspired by the representation learning pre-training methods, Our GPS adopts the supervised contrastive loss to calculate gradient (without random initialization of the classification head). As shown in \cref{tab:ablation} (c), when we use the cross-entropy loss (with the head) to calculate the gradient, the average accuracy on FGVC is dropped by 0.67\%, illustrating the importance of obtaining accurate gradients.

\paragraph{Selected connection number} 
As shown in \cref{fig:ablation:Figure_Ablation_Num}
we select top-K input connections per neuron as trainable parameters, ranging from 1 to 15, and conduct experiments on the 5 tasks. We can observe that more trainable parameters do not necessarily lead to better performance, but each data set has a performance peak. In addition, on the dataset with sufficient training data, the addition of trainable parameters can greatly improve the accuracy. Our GPS can easily control the number of trainable parameters and achieve optimal results on each dataset.

\paragraph{Robustness to seeds} Addition-based fine-tuning methods like VPT are sensitive to the initialization of additional parameters as well as random seeds, whereas select-based methods are not. All results in \cref{tab:ablation} are the average accuracy of three seeds on FGVC datasets (Only shows the std of GPS here. Please see details in supplementary). The results show random seeds have little influence on our method.

\section{Conclusion}
\label{conclusion}
In this paper, we propose a new paradigm for PEFT, $i.e.$ Gradient-based Parameter Selection (GPS). Our approach does not introduce any additional parameters and only fine-tunes a small subset of the pre-trained model's parameters for downstream tasks, resulting in robust generalization across diverse models and adaptively selecting a subset of parameters for each task. 
Remarkably, GPS achieves significant 
improvement on a range of tasks (including image classification and semantic segmentation), compared to the full fine-tuning method.
GPS also attains SOTA performance compared to other PEFT methods.

\section*{Acknowledgement}
Shanghang Zhang is supported by the National Science and Technology Major Project of China (No. 2022ZD0117801).

{
    \small
    \bibliographystyle{ieeenat_fullname}
    \bibliography{main}
}

\appendix
\clearpage
\setcounter{page}{1}
\maketitlesupplementary

\section{Details of experiments}
\label{Details of experiments}
\subsection{Baseline description}

\paragraph{Vision Transformer (ViT)} As a transformer-based visual model, ViT~\cite{dosovitskiy2020image} has been widely adopted in various visual tasks. Most of the experiments are conducted on pre-trained ViT architecture in this paper. Given an input image $I \in \mathbb{R}^{H \times W \times 3}$, before feeding the image into the Transformer, the image is partitioned into $M$ patches and appended a [CLS] token for classification purposes, resulting in final input $x \in \mathbb{R}^{ (M+1) \times d}$ where $d$ is the dimension of the features. The Transformer typically consists of multiple blocks and each block contains a Multi-head Attention layer (MHA) and two MLP layers\cite{vaswani2017attention}.

\paragraph{Adapter} Work in~\cite{houlsby2019parameter} proposed the Adapter method, which inserts multiple trainable layers (termed as Adapter) into the pre-trained Transformer encoder. Only the Adapter is updated during the fine-tuning stage. These layers can be inserted after either the Multi-head Attention layer or the MLP layer. Adapter comprises two projection matrices, one $W^{\text {down}}$ for dimension reduction and the other $W^{\text {up}}$ for feature reconstruction to the original dimension. Specifically, given the input $x \in \mathbb{R}^{ (M+1) \times d}$, the output of the Adapter is 
\begin{equation} 
y =\left[W^{\text {up }} \phi\left(W^{\text {down }} x^{T}\right)\right]^{T}
\end{equation}
where $W^{\text {up }} \in \mathbb{R}^{d^{\prime} \times d}$, $W^{\text {down }} \in \mathbb{R}^{d \times d^{\prime}}$ (where $d^{\prime} \ll d$ ), and $\phi$ is a nonlinear activation function.

\paragraph{Prompt} Visual prompt tuning (VPT) introduces learnable parameters ($i.e.$, prompts) into the input space~\cite{jia2022visual}. When fine-tuning downstream tasks, the backbone is fixed, and just tuning these prompts. Formally, the given input $x \in \mathbb{R}^{ (M+1) \times d}$ is concatenated with $m$ introduced prompts $p \in \mathbb{R}^{m \times d}$. The final combined input is
\begin{equation} 
x^{\prime}=[x ; p]
\end{equation}
where $x^{\prime} \in \mathbb{R}^{\left(M+1+m\right) \times d}$ will be feed into the Transformer. There are two versions of VPT, namely VPT-shallow and VPT-deep. The former introduces learnable prompts solely into the input space of the first layer, whereas the latter integrates them into each layer's input space.

\paragraph{Scale and shift feature} SSF attempts to scale and shift the features between the layers of the pre-trained model by adding a linear transform layer~\cite{lian2022scaling}. During fine-tuning the downstream tasks, only the linear transform layers are updated while the backbone remains frozen. The transform layer consists of two components, scale factor $\gamma \in \mathbb{R}^{d}$ and shift factor $\beta \in \mathbb{R}^{d}$, for feature transformation. To be specific, given the input $x \in \mathbb{R}^{\left(M+1\right) \times d}$,  the output is calculated by
\begin{equation}
y=\gamma \odot x+\beta
\end{equation}
where  $y \in \mathbb{R}^{\left(M+1\right) \times d}$,  $\odot$ is the dot product.

\begin{figure}[!t]
    \centering
    \subfigure[One input connection]{		\includegraphics[width=0.46\linewidth]{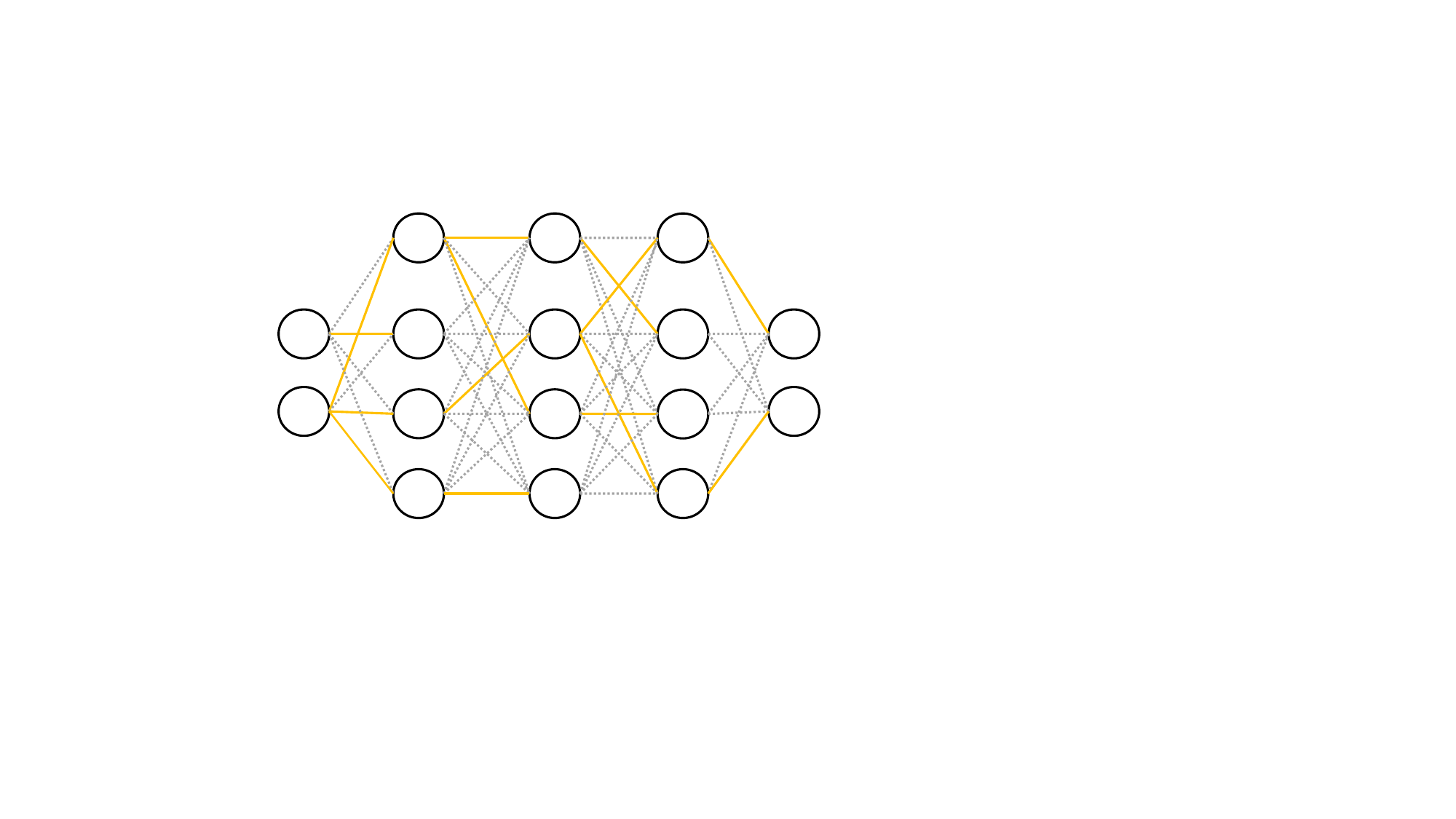}
    \label{fig:connection number_one}
    }
    \subfigure[Two input connections]{		\includegraphics[width=0.46\linewidth]{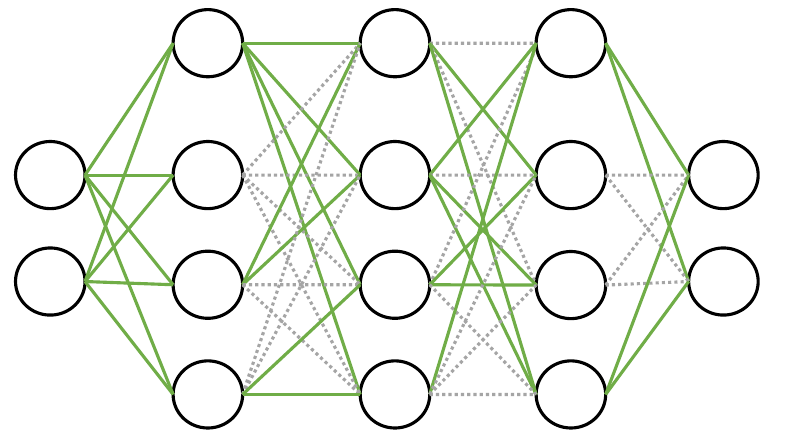}
    \label{fig:connection number_two}
    }
    \caption{Different number of connections with highest gradient value among all input connections per neuron, such as (a) selecting only one input connection per neuron. (b) two input connections are selected per neuron.}
    \label{fig:connection number}
\end{figure}

\subsection{The number of parameters on different tasks}

\begin{table*}[!t]
\centering
\resizebox{1\linewidth}{!}{
\begin{tabular}{lcc|lcc|lcc} 
\toprule
		Dataset        & Params. (M) & Conne.s & Dataset & Params. (M) & Conne.s & Dataset & Params. (M) & Conne.s \\ \midrule
		CUB-200-2011   & 0.47      &   2    & Pets      & 0.23      & 1      &DMLab       & 0.20      &1  \\
		NABirds        &   1.35    & 10      & SVHN      & 0.20      & 1      &KITTI/distance       &0.20       &1  \\
		Oxford Flowers  & 0.29      &   1    & Sun397      &  0.55     &  1     &dSprites/loc        &0.21       &1  \\
		Stanford Dogs  & 0.30      &  1     & Patch Camelyon      & 0.30      & 2      &dSprites/ori      &0.21      &1  \\
		Stanford Cars  & 1.07      & 10      &  EuroSAT     & 0.20      &  1     &SmallNORB/azi       &0.21       & 1 \\
		$\text{CIFAR-100}^\star$ & 0.29      &  1     &  Resisc45     &0.24       & 1      &SmallNORB/ele        &0.20       &1  \\
		Caltech101      &  0.29     & 1      &Retinopathy       & 0.20      &1      &  CIFAR-100     &   0.58   &   5\\
	    DTD            & 0.24      & 1      & Clevr/count      &0.30       &1       &     CIFAR-100 (Swin)  &   0.82    &  5\\
		Flowers102     & 0.29      & 1      & Clevr/distance      & 0.20      &1       &  CIFAR-100 (ConvNeXt)     &   0.78    &  5 \\ \bottomrule 
\end{tabular}}
\caption{The number of learnable parameters and connections across all tasks. $\text{CIFAR-100}^\star$ is a subset of CIFAR-100 in VTAB benchmark. In bracket is the model architecture, without bracket represents the one fine-tuned on ViT-B/16. Params. means the learnable parameters and the Conne. represents the number of selected input connections for each neuron in the network.} 
\label{tabel: number of parameter}
\end{table*}

\begin{figure}[!t]
    \centering
    \includegraphics[width=0.8\linewidth]{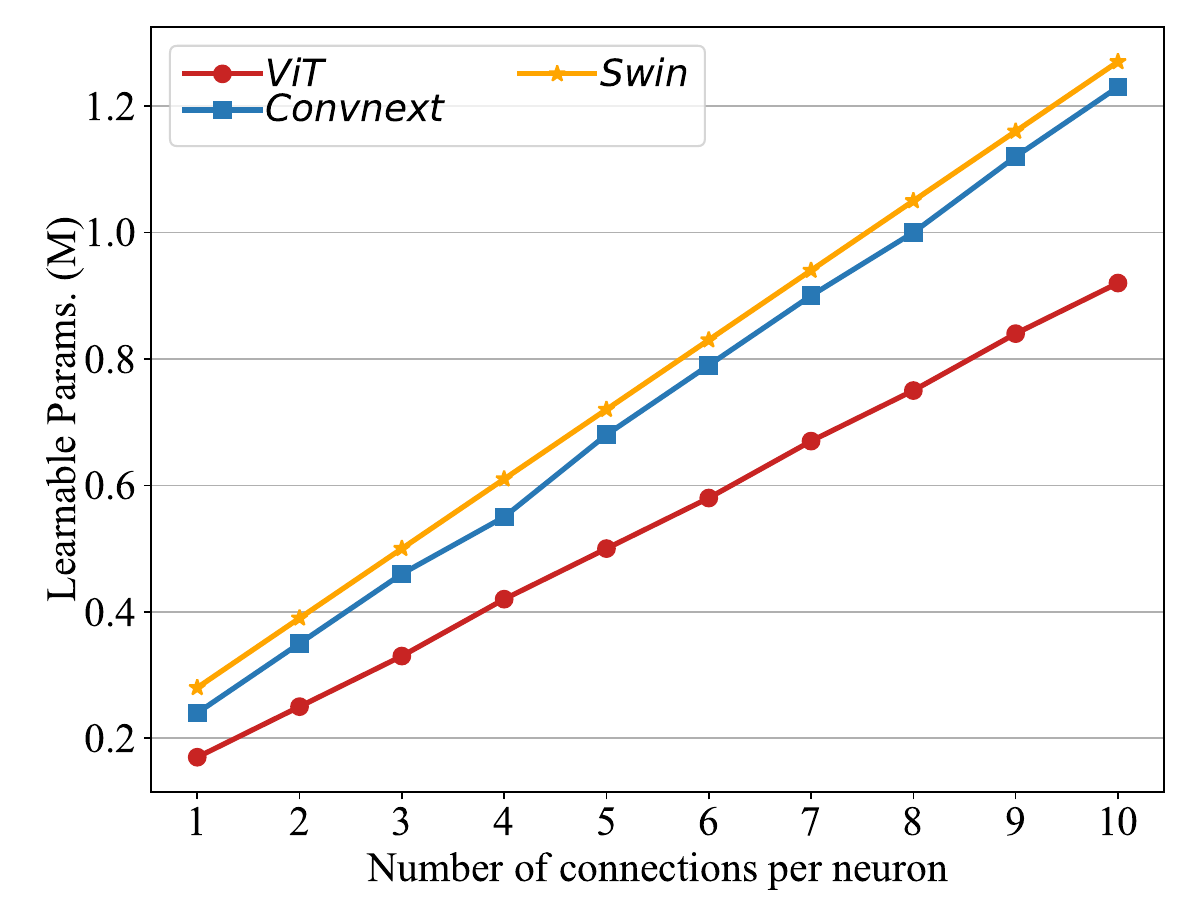}
    \caption{The number of learnable parameters with the different number of connections on VIT-B/16, Swin and Convnext archite. The learnable parameters do not contain the task-specific head.}
    \label{fig:number of parameters with connection}
\end{figure}

For each neuron in the network, our GPS method selected at least one of the connections (weight or parameter) with the highest gradient value, among the input connections of the neuron, as shown in~\cref{fig:connection number_one}. For downstream tasks that need more learnable parameters to better fit the data, such as  those tasks with dissimilar data distributions from the upstream dataset (such as NABirds) or larger amounts of data (such as CIFAR-100), our method can be easily extended by introducing more learnable parameters. Specifically, for each neuron, we can select multiple input connections with the highest gradient values instead of limiting them to just one, as shown in~\cref{fig:connection number_two}. \cref{tabel: number of parameter} show the detailed statics on the number of parameters that are selected in our paper. For most of the tasks in this paper, we just select one of the connections. We also explore the relationship between the number of connections and the number of learnable parameters. As shown in~\cref{fig:number of parameters with connection}, with the increase in the number of selected connections with the highest gradient value among the input connections per neuron, the number of learnable parameters linear ascent.

\begin{figure*}[tpbh]
    \centering

    \subfigure[CUB-200-2011]{		\includegraphics[width=0.31\linewidth]{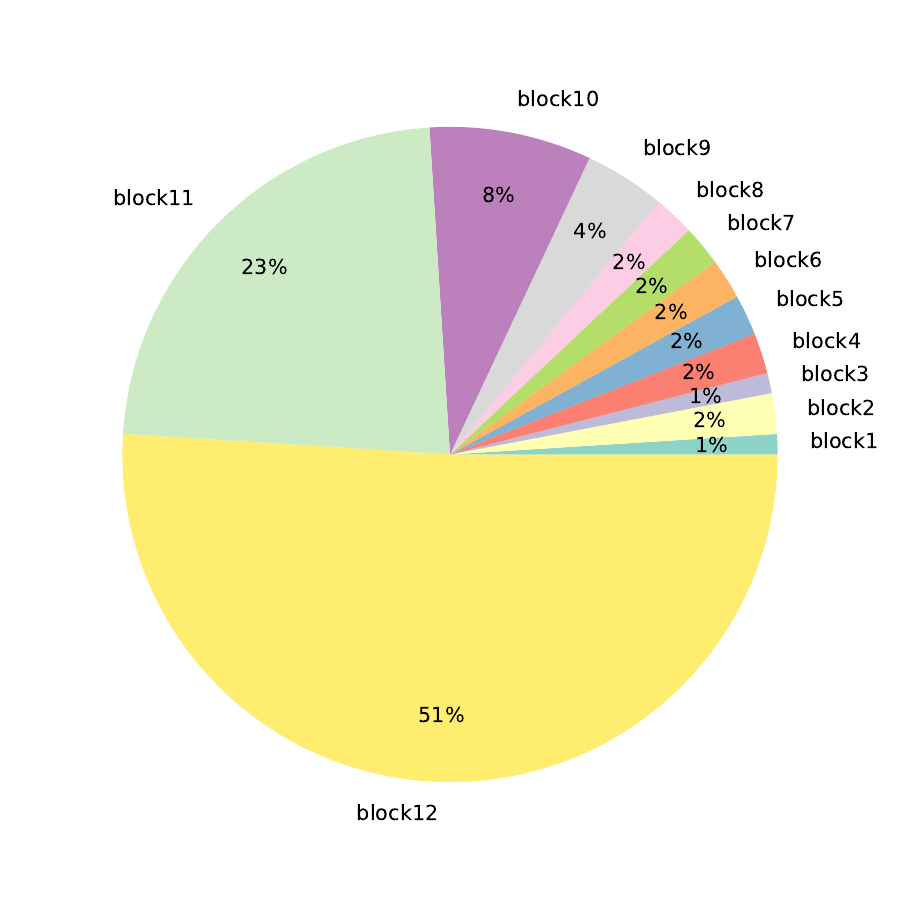}
    \label{fig: NET selection_cub}
    }
    \subfigure[Oxford Flowers]{		\includegraphics[width=0.31\linewidth]{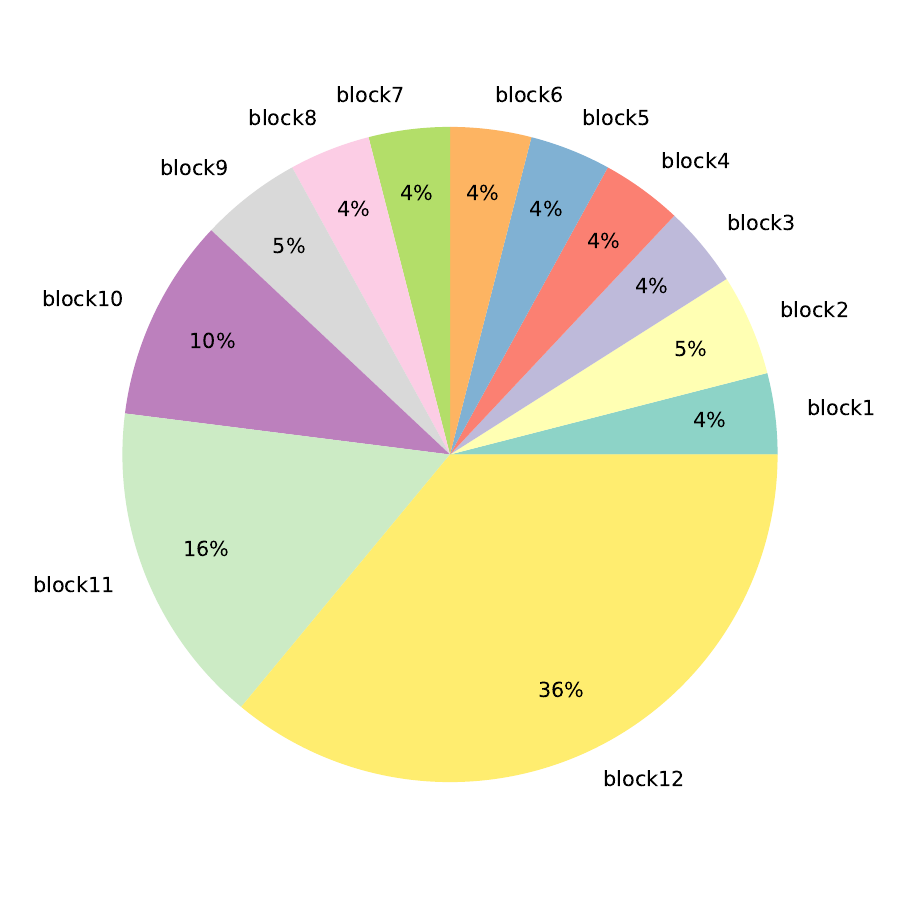}
    \label{fig: NET selection_flower}
    }
    \subfigure[Stanford Dogs]{		\includegraphics[width=0.31\linewidth]{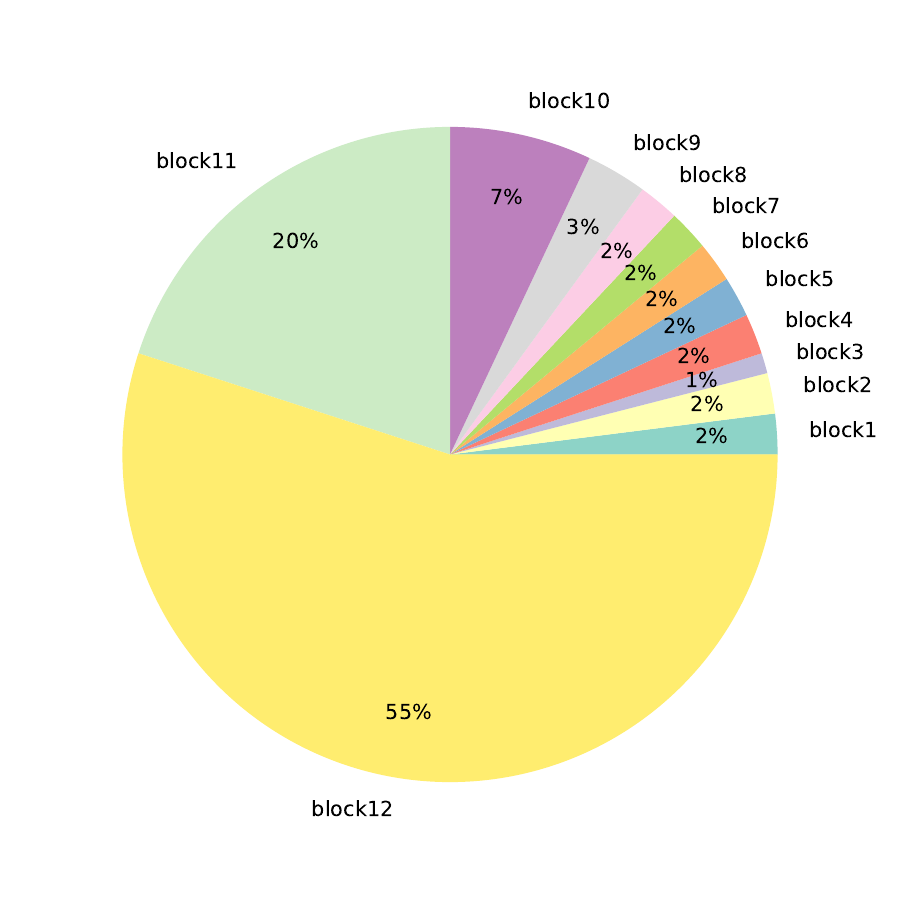}
    \label{fig: NET selection_dog}
    }
    \subfigure[Stanford Cars]{		\includegraphics[width=0.31\linewidth]{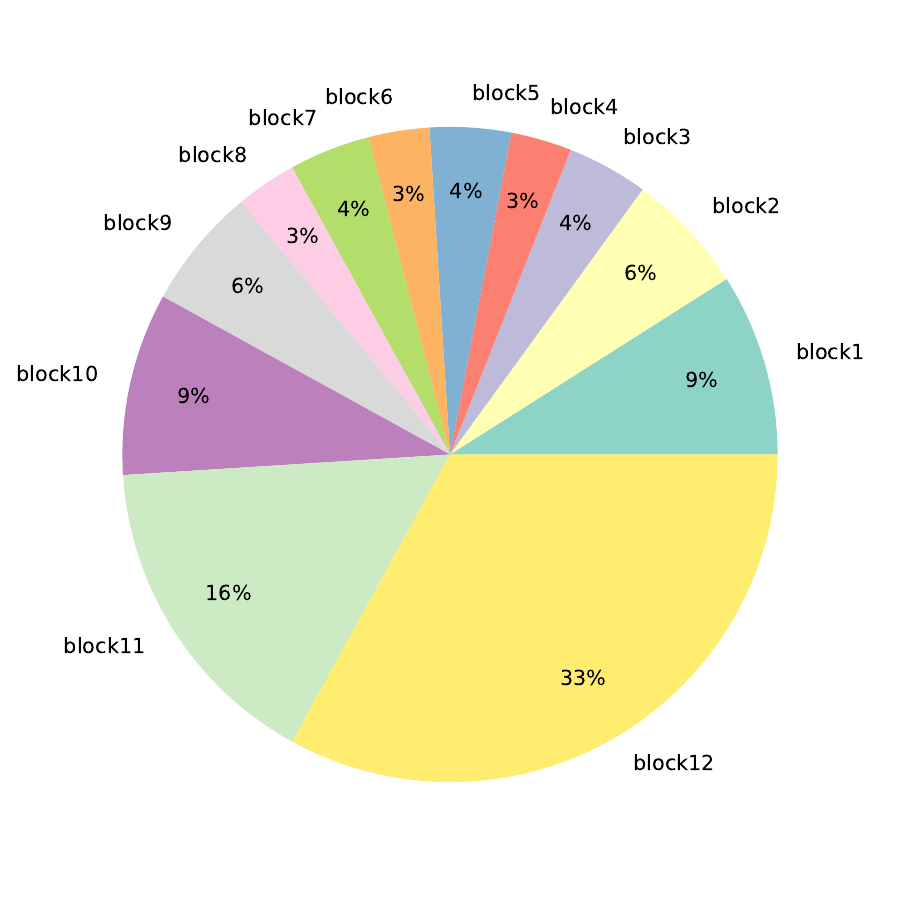}
    \label{fig: NET selection_car}
    }
    \subfigure[NABirds]{		\includegraphics[width=0.31\linewidth]{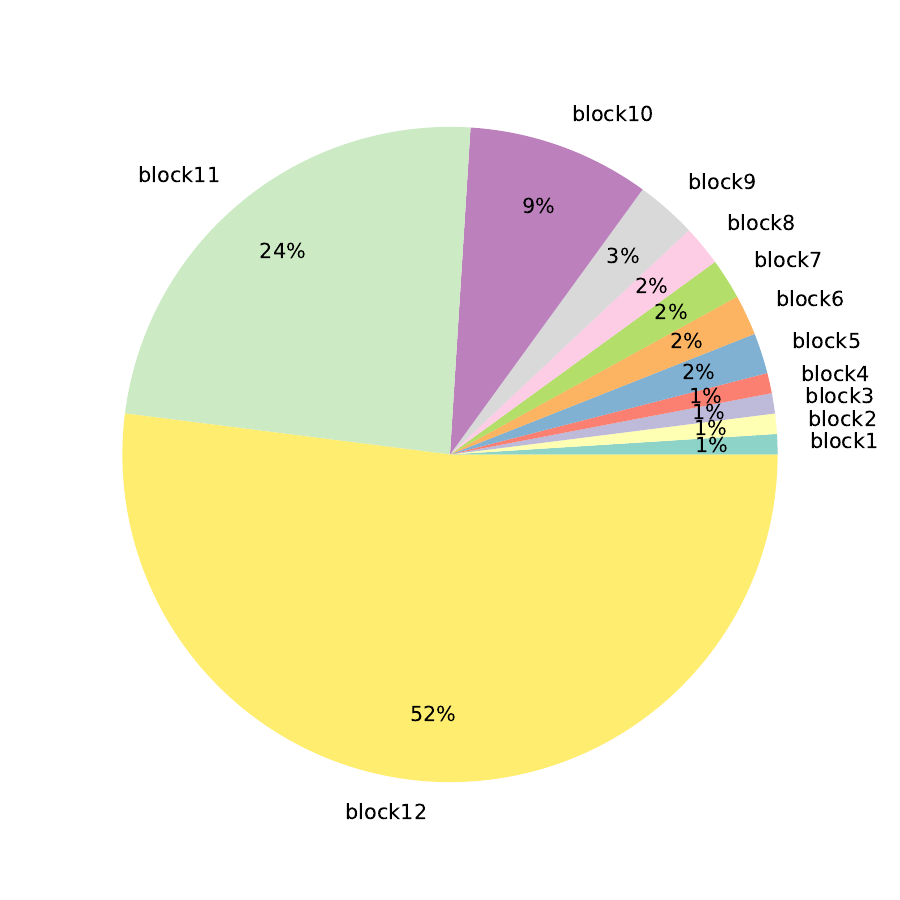}
    \label{fig: NET selection_nabird}
    }
    \subfigure[GPS]{		\includegraphics[width=0.31\linewidth]{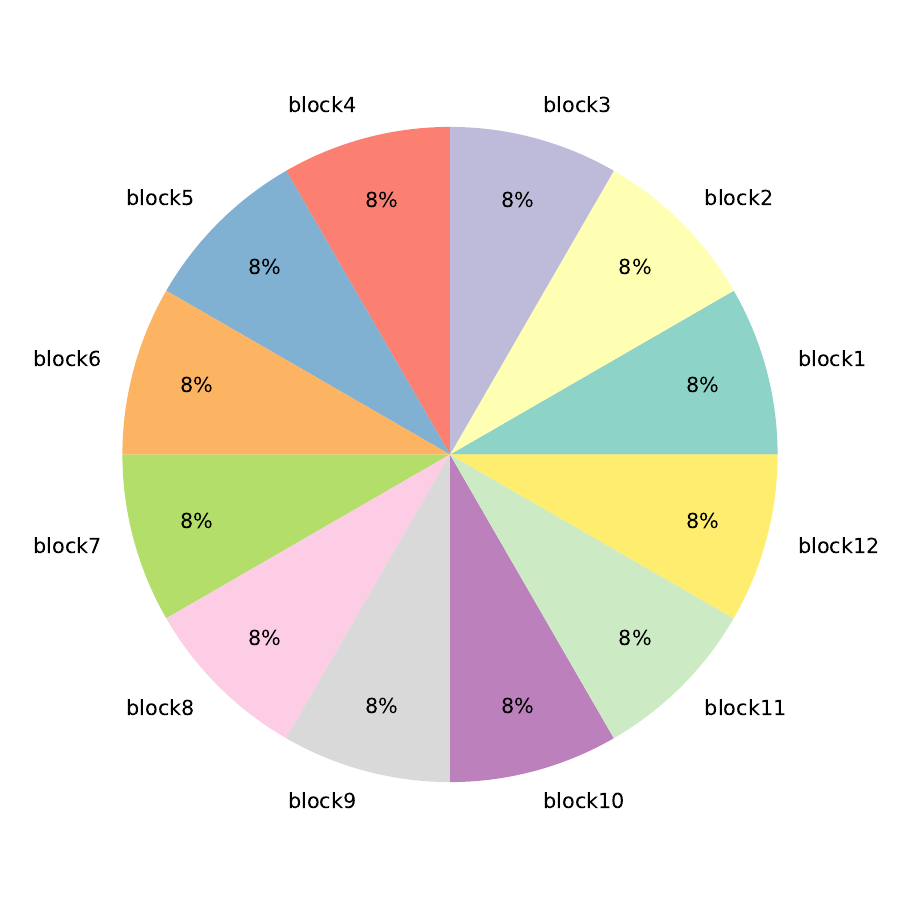}
    \label{fig: NET selection_gps}
    }
    \caption{Distribution of the parameter over the whole network ViT-B/16 on 6 FGVC dataset (CUB-200-2011,  Oxford Flowers, Stanford Dogs, Stanford Cars and NABirds). Selecting the 1\% parameters with the highest gradient value from the whole network, instead of our method selecting at least one of the connections among all input connections per neuron. In contrast to this method, our GPS has the same distribution over different downstream tasks (f).}
    \label{fig: NET selection}
\end{figure*}

\subsection{Parameters distribution of \textit{Net} selection}
In contrast to our approach, a simple approach is to
select the parameters for a specific task by selecting a certain percentage of parameters with the highest gradient from the entire network~\cite{he2023sensitivity}. However, as shown in~\cref{fig: NET selection_cub,fig: NET selection_flower,fig: NET selection_dog,fig: NET selection_car,fig: NET selection_nabird}, most of the selected parameters are located in the upper layers, specifically block 12 and block 11. As a result, the network is primarily focused on fine-tuning abstract features while lacking the ability to fine-tune detailed information from shallower layers. Our approach addresses this challenge by carefully selecting the input connections for each individual neuron, resulting in our selected parameters being evenly distributed on the whole network, as shown in~\cref{fig: NET selection_gps}.

\section{Additional experiments}
\label{Additional experiments}
\subsection{Robustness and OOD datasets}
\label{Robustness and OOD Datasets}

\begin{table}
    \centering
    \resizebox{1\linewidth}{!}{
        \begin{tabular}{c|c|c|c|c}
        \toprule
        \diagbox{Method}{Dataset} & \makecell[c]{ImageNet\\-1K ($\uparrow$)} & \makecell[c]{ImageNet\\-A ($\uparrow$)} & \makecell[c]{ImageNet\\-R ($\uparrow$)} & \makecell[c]{ImageNet\\-C ($\downarrow$)} \\
        \midrule
        Full~\cite{jia2022visual}                                 & 83.58         & 34.49       & 51.29    &  46.47     \\
        \midrule
        Linear~\cite{jia2022visual}                               & 82.04       & 33.91      & 52.87  & 46.91   \\
        Bias~\cite{zaken2021bitfit}                                 & 82.74       & 42.12      & 55.94   & 41.90     \\
        \midrule
        Adapter~\cite{houlsby2019parameter}        & 82.72           & 42.21      & 54.13   & 42.65     \\
        VPT-Shallow~\cite{jia2022visual}                          & 82.08             & 30.93      & 53.72   & 46.88        \\
        VPT-Deep~\cite{jia2022visual}                             & 82.45           & 39.10      & 53.54   & 43.10    \\
        SSF~\cite{lian2022scaling}                                  & 83.10                & 45.88       & 56.77     &    \textbf{41.47}    \\
        \midrule
        GPS                             &   \textbf{83.91}             & \textbf{46.11}      & \textbf{57.00}    & 42.04         \\
        
        \bottomrule
        \end{tabular}
    }
    \caption{Performance comparisons on the ImageNet with different model architectures.}
    \label{tab:ood}
\end{table}

In addition to standard classification tasks, we further analyze the robustness and OOD generalization ability of GPS. Based on the Imagenet-A, ImageNet-R, and ImageNet-C datasets, we first fine-tune the model on ImageNet-1K, and then test the fine-tuned model on the three datasets respectively. The results are shown in ~\cref{tab:ood}. GPS not only achieves the best performance on the standard ImageNet-1K classification task but also achieves good performance in robustness and generalization tests. Among them, GPS achieves the best results on ImageNet-A and ImageNet-R, outperforms the previous optimal SSF by 0.23\%, reflecting the strong stability and generalization  ability of our method. On ImageNet-C, GPS performs slightly worse, lagging behind SSF, but still higher than addition-based Adapter and VPT. This result indicates that our method can quickly adapt to the data distribution of downstream tasks, but it needs to be improved in anti-interference.

\begin{table*}[!t]
    \centering
    \resizebox{1\linewidth}{!}{
        \begin{tabular}{c|ccccc|ccc}
        \toprule
        \diagbox{Method}{Dataset} & \makecell[c]{CUB-200\\-2011} & NABrids & \makecell[c]{Oxford\\Flowers} & \makecell[c]{Stanford\\Dogs} & \makecell[c]{Stanford\\Cars} & \makecell[c]{Mean\\Acc.} & \makecell[c]{Mean\\Params. (M)} & \makecell[c]{Mean\\Params. (\%)} \\
        \midrule
        ViT-B/16 + Full           &  87.3 & 82.7 & 98.8 & 89.4 & 84.5 & 88.54 & 85.98 & 100.00 \\
        ViT-B/16 + Linear         &  85.3 & 75.9 & 97.9 & 86.2 & 51.3 & 79.32 & 0.18 & 0.21 \\
        ViT-B/16 + SSF            &  89.5 & 85.7 & 99.6 & 89.6 & 89.2 & 90.72 & 0.39 & 0.45 \\
        ViT-B/16 + GPS (Ours)     & \textbf{89.9}         & \textbf{86.7}    & \textbf{99.7}           & \textbf{92.2}          & \textbf{90.4}          & \textbf{91.78}     & 0.66 & 0.77             \\
        \midrule
        Swin-B + Full           &  90.7 & \textbf{89.8} & 99.5 & 88.9 & \textbf{93.2} & 92.42 & 86.98 & 100.00 \\
        Swin-B + Linear         &  90.6 & 86.8 & 99.2 & 88.3 & 74.6 & 87.90 & 0.24 & 0.28 \\
        Swin-B + SSF            &  90.5 & 88.4 & \textbf{99.7} & 88.7 & 90.4 & 91.54 & 0.49 & 0.56 \\
        Swin-B + GPS (Ours)     & \textbf{90.8}         & 88.9    & \textbf{99.7}           & \textbf{92.7}         & 90.7          & \textbf{92.56}     & 0.83 & 0.95             \\
        \midrule
        ConvNeXt-B + Full           &  \textbf{91.2} & \textbf{90.4} & 99.6 & 89.9 & \textbf{94.1} & 93.04 & 87.81 & 100.00 \\
        ConvNeXt-B + Linear         &  90.6 & 86.9 & 99.3 & 89.7 & 73.5 & 88.00 & 0.24 & 0.28 \\
        ConvNeXt-B + SSF            &  90.8 & 89.0 & \textbf{99.7} & 90.4 & 92.5 & 92.48 & 0.50 & 0.56 \\
        ConvNeXt-B + GPS (Ours)     & 91.0         & 89.6    & \textbf{99.7}           & \textbf{93.7}          & 92.6          & \textbf{93.32}     & 0.79 & 0.90             \\
        \bottomrule
        \end{tabular}
    }
    \caption{Performance comparisons on FGVC benchmark with different model architectures.}
    \label{tab:fgvc_arch_all}
\end{table*}

\subsection{More experiments on different architecture}

 As mentioned in the main body of our paper, our method is model-agnostic, we further compare GPS with other fine-tuning methods across ViT-B/16, Swin-B, and ConvNeXt-B architectures on the ImageNet-1k~\cite{deng2009imagenet} and CIFAR-100~\cite{krizhevsky2009learning} datasets.

\paragraph{CIFAR-100} As shown in \cref{tab:cifar}, unlike FGVC and VTAB, GPS and other efficient tuning methods have difficulty in achieving competitive performance as full tuning on CIFAR-100. This may be due to that CIFAR-100 contains more training data, allowing all parameters of the entire model to be adequately trained, which seriously reduces the advantages of efficient fine-tuning methods. However, GPS still outperforms all previous parameter-efficient tuning methods (Bias, Adapter, VPT, and SSF) and reduces the gap with full fine-tuning to less than 0.5\% on all architectures, which further demonstrates the adaptability of our approach to different models.

\paragraph{ImageNet-1k} Similar to the results on CIFAR-100, ImageNet-1K contains more training data, which makes it harder for parameter-efficient fine-tuning algorithms to achieve the same accuracy as full fine-tuning, as shown in~\cref{tab:diff_struct}. However, GPS still outperforms full fine-tuning by 0.33\% on ViT structure, and outperforms the previous SOTA method SSF on Swin and ConvNeXt structures respectively, which further shows the generalization of GPS to different model structures.

\begin{table}[!t]
    \centering
    \resizebox{1\linewidth}{!}{
        \begin{tabular}{c|cc|cc|cc}
        \toprule
        \multirow{2}{*}{Architecture} & \multicolumn{2}{c|}{ViT-B/16} & \multicolumn{2}{c|}{Swin-B} & \multicolumn{2}{c}{ConvNeXt-B} \\
        \cmidrule{2-7}
                                             & Acc.       & Params.(\%)      & Acc.      & Params.(\%)     & Acc.        & Params.(\%)       \\
        \midrule
        Full~\cite{jia2022visual}                                 & 93.82      & 100.00          & \textbf{93.85}     & 100.00         & \textbf{94.14}       & 100.00           \\
        \midrule
        Linear~\cite{jia2022visual}                     & 88.70      & 0.09            & 89.27     & 0.12           & 89.20       & 0.12             \\
        Bias~\cite{zaken2021bitfit}          & 93.39      & 0.21            & 92.19     & 0.28           & 92.80       & 0.27             \\
        \midrule
        Adapter~\cite{houlsby2019parameter}     & 93.34      & 0.36            & 92.49     & 0.38           & 92.86       & 0.52             \\
        VPT-Shallow~\cite{jia2022visual}   & 90.38      & 1.07            & 90.02     & 0.15           & -           & -                \\
        VPT-Deep~\cite{jia2022visual}      & 93.17      & 1.43            & 92.62     & 0.81           & -           & -                \\
        SSF~\cite{lian2022scaling}    & \underline{93.99}      & 0.33            & 93.06     & 0.43           & 93.45       & 0.42             \\
        \midrule
        GPS (Ours)                             & \textbf{94.02}      & 0.68            & \underline{93.55}     & 0.96           & \underline{93.58}       & 0.90             \\
        \bottomrule
        \end{tabular}
    }    
    \caption{Performance comparisons on the CIFAR-100 with different model architectures.}
    \label{tab:cifar}
\end{table}

\begin{table}[!t]
    \centering
    \resizebox{1\linewidth}{!}{
        \begin{tabular}{c|cc|cc|cc}
        \toprule
        \multirow{2}{*}{Architecture} & \multicolumn{2}{c|}{ViT-B/16} & \multicolumn{2}{c|}{Swin-B} & \multicolumn{2}{c}{ConvNeXt-B} \\
        \cmidrule{2-7}
                                             & Acc.       & Params.(\%)      & Acc.      & Params.(\%)     & Acc.        & Params.(\%)       \\
        \midrule
        Full~\cite{jia2022visual}                                 & \underline{83.58}      & 100.00          & \textbf{85.20}    & 100.00         & \textbf{85.80}       & 100.00           \\
        \midrule
        Linear~\cite{jia2022visual}                               & 82.04      & 0.89            & 83.25     & 1.17           & 84.05       & 1.16             \\
        Bias~\cite{zaken2021bitfit} & 82.74      & 1.00            & 83.92     & 1.32           & 84.63       & 1.31             \\
        \midrule
        Adapter~\cite{houlsby2019parameter}& 82.72      & 1.16           & 83.82    & 1.43           & 84.49      & 1.54            \\
        VPT-Shallow~\cite{jia2022visual}                          & 82.08      & 1.06            & 83.29     & 1.19           & -           & -                \\
        VPT-Deep~\cite{jia2022visual}                             & 82.45      & 1.42            & 83.44     & 1.85           & -           & -                \\
        SSF~\cite{lian2022scaling}                                   & 83.10     & 1.12            & 84.40     & 1.47           & 84.85       & 1.44             \\
        \midrule
        GPS (ours)                            & \textbf{83.91}      & 1.37            &  \underline{84.43}     & 1.96       &  \underline{84.87}      & 1.90   \\
        \bottomrule
        \end{tabular}
    }
    \caption{Performance comparisons on the ImageNet-1k with different model architectures.}
    \label{tab:diff_struct}
\end{table}

\paragraph{FGVC} As mentioned in the main body of our paper, our method achieves the best result on FGVC benchmark. \cref{tab:fgvc_arch_all} shows the full results of \cref{tab:architecture} in the main body. Among all three model architectures, GPS consistently outperforms all other baselines, demonstrating its model-agnostic advantage.

\subsection{Data-efficient tuning}
\label{Data-efficient tuning}

\begin{figure}
    \centering
    \includegraphics[width=0.8\linewidth]{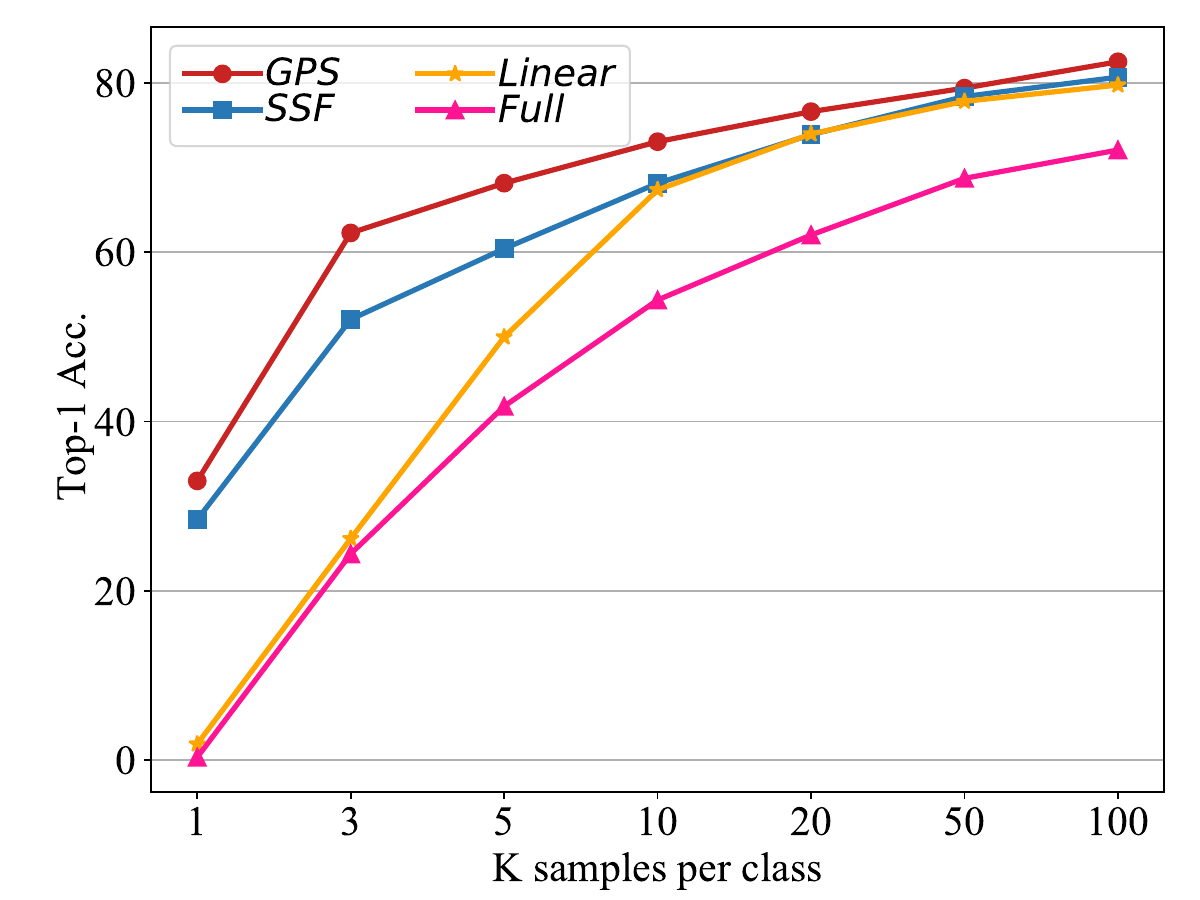}
    \caption{The performance comparison of different fine-tuning methods under k values for each class on ImageNet dataset. Our method is always above the curve of the others. The advantage of our approach is particularly evident in extreme cases where data is extremely scarce (e.g., k=1). }
    \label{fig:fewshot}
\end{figure}

Recent advances in large foundation model fine-tuning have shown considerable promise in reaching state-of-the-art performance on various tasks. However, in order to reach high accuracy, these methods often need significant volumes of training data, which may be time-consuming and costly to obtain. Here we demonstrate that our method is data efficient, that is, with such a few-shot setting, our method requires only a small amount of training data for tuning to achieve outstanding results that other approaches do not. Specifically, we fine-tune the ViT-B/16 by selecting only k samples for each class in the ImageNet dataset to form a few-shot training set. The value of k and the accuracy of predicted results are illustrated in~\cref{fig:fewshot}, which demonstrates the excellent data efficiency of our method especially in extreme cases like k=1.

\subsection{Random seed for impacts of different selection schemes and ablations}
We conduct experiments with three random seeds to investigate the robustness of our method. As shown in~\cref{tab:appendix: ablation}, Our parameter selection method significantly outperforms the other methods with small randomness.~\cref{tab:appendix: ablation} is a supplementary of ~\cref{tab:ablation} in the main body of our paper.

\begin{table*}[!t]
\centering
\resizebox{0.7\linewidth}{!}{
\begin{tabular}{ll|lllll}
\toprule
 &             & CUB   & NAbirds & Flowers & Cars  & Dogs  \\ \midrule
\multicolumn{1}{l|}{\multirow{2}{*}{(a)}} & Net            & 86.86 ± 0.21     & 86.55 ± 0.03    & 99.62 ± 0.01   & 89.65 ± 0.12 & 91.32 ± 0.07   \\
\multicolumn{1}{l|}{}                     & Layer          & 87.30 ± 0.13 & 86.79 ± 0.08   & 99.64 ± 0.01   & 90.03 ± 0.13 & 91.90 ± 0.11 \\ \midrule
\multicolumn{1}{l|}{\multirow{3}{*}{(b)}} & Net Random     & 86.60 ± 0.10 & 85.98 ± 0.07   & 99.61 ± 0.01    & 89.10 ± 0.12 & 91.34 ± 0.12 \\
\multicolumn{1}{l|}{}                     & Neuron Random  & 87.17 ± 0.15 & 86.02 ± 0.10   & 99.62 ± 0.01   & 89.52 ± 0.23 & 91.82 ± 0.23 \\
\multicolumn{1}{l|}{}                     & Magnitude      & 87.29 ± 0.12 & 85.99 ± 0.08   & 99.62 ± 0.00   & 89.29 ± 0.02  & 91.30 ± 0.02  \\ \midrule
\multicolumn{1}{l|}{(c)}                  & Head+CE        & 87.05 ± 0.19  & 86.20 ± 0.14   & 99.64 ± 0.01   & 89.25 ± 0.09 & 91.29 ± 0.01  \\ \midrule
&GPS           & \textbf{88.07} ± 0.11  & \textbf{86.64} ± 0.03 & \textbf{99.69} ± 0.01 & \textbf{90.10} ± 0.10 & \textbf{92.30} ± 0.10  \\ \bottomrule               
\end{tabular}}
\caption{Impacts of different selection schemes and ablations. (a) Different selection levels. (b) Different selection criteria. (c) Different ways to calculate gradients.}
\label{tab:appendix: ablation}
\end{table*}

\begin{figure*}[!t]
    \centering
    \includegraphics[width=0.8\linewidth]{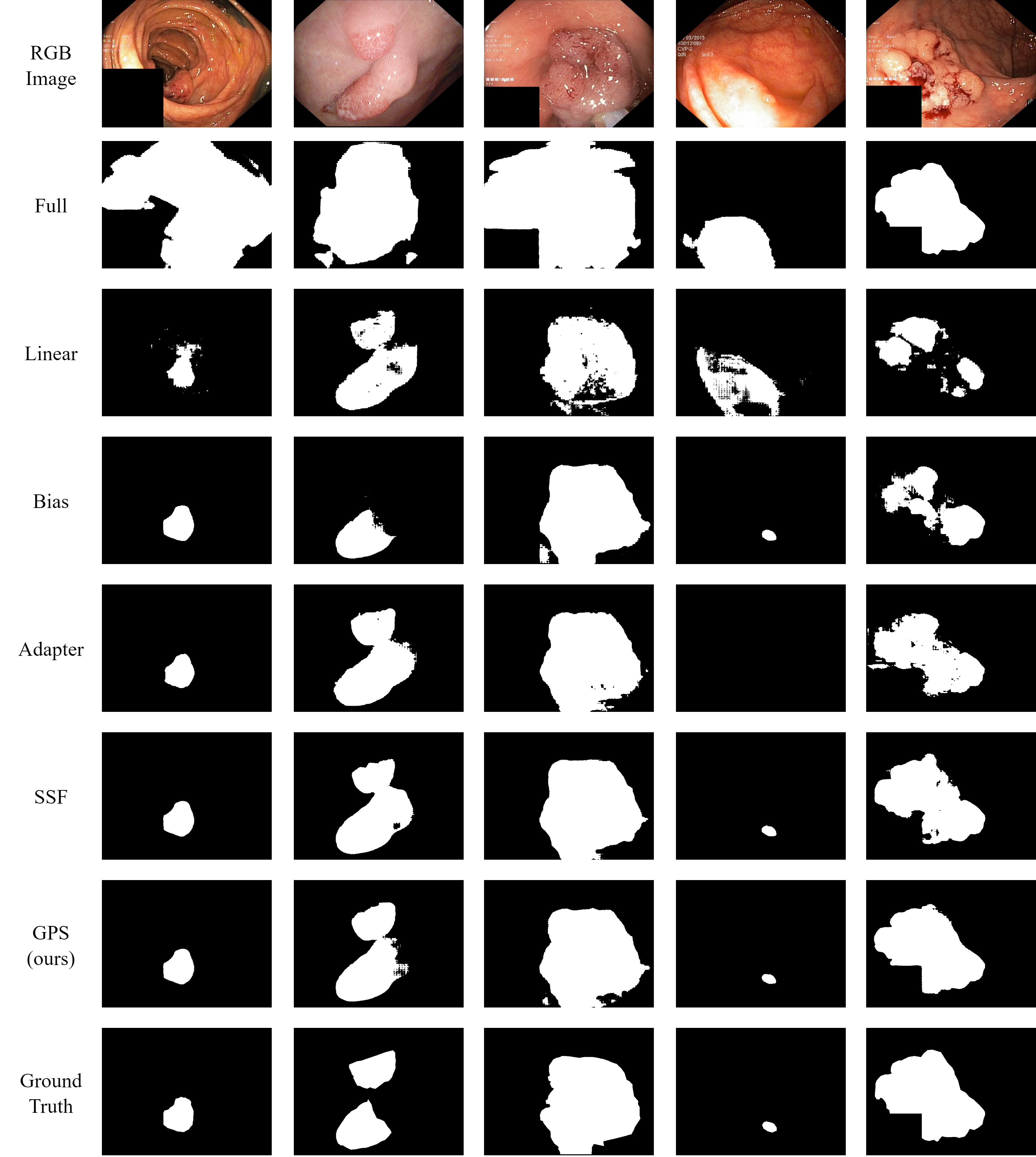}
\caption{\textbf{The Visualization Result of Polyp Segmentation.} Here GT means ground truth. As illustrated in this figure, although SAM and other methods can identify some polyp structures in the image, the result is not accurate. By using GPS, our approach elevates the performance with SAM.}
    \label{fig:polyp}
\end{figure*}

 \begin{figure*}[!t]
    \centering
    \subfigure[]{		\includegraphics[width=0.23\linewidth]{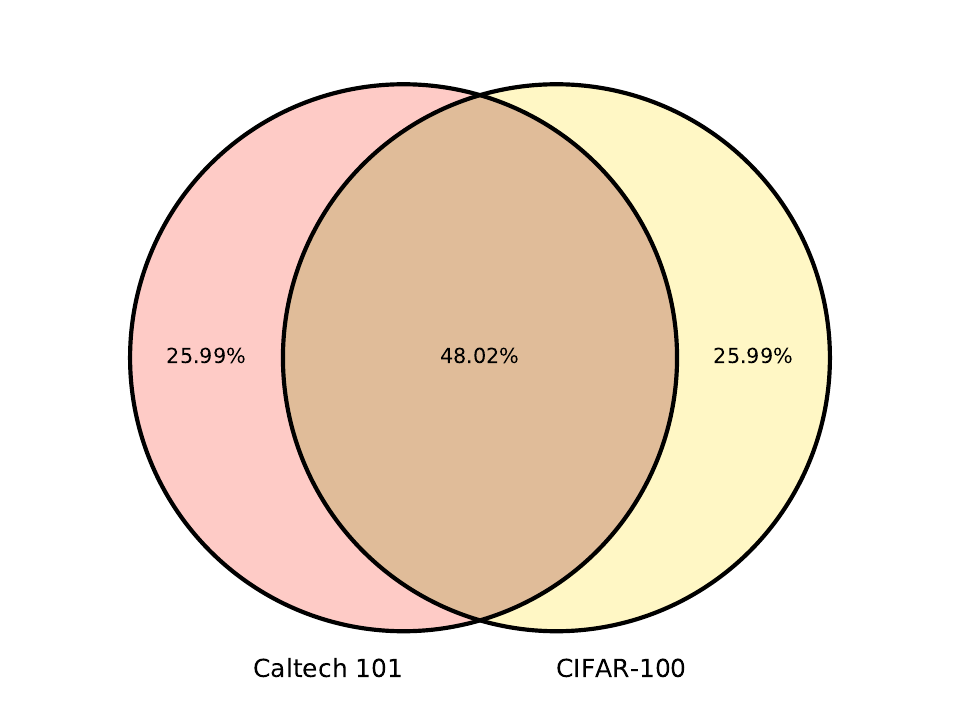}
    }
    \subfigure[]{		\includegraphics[width=0.23\linewidth]{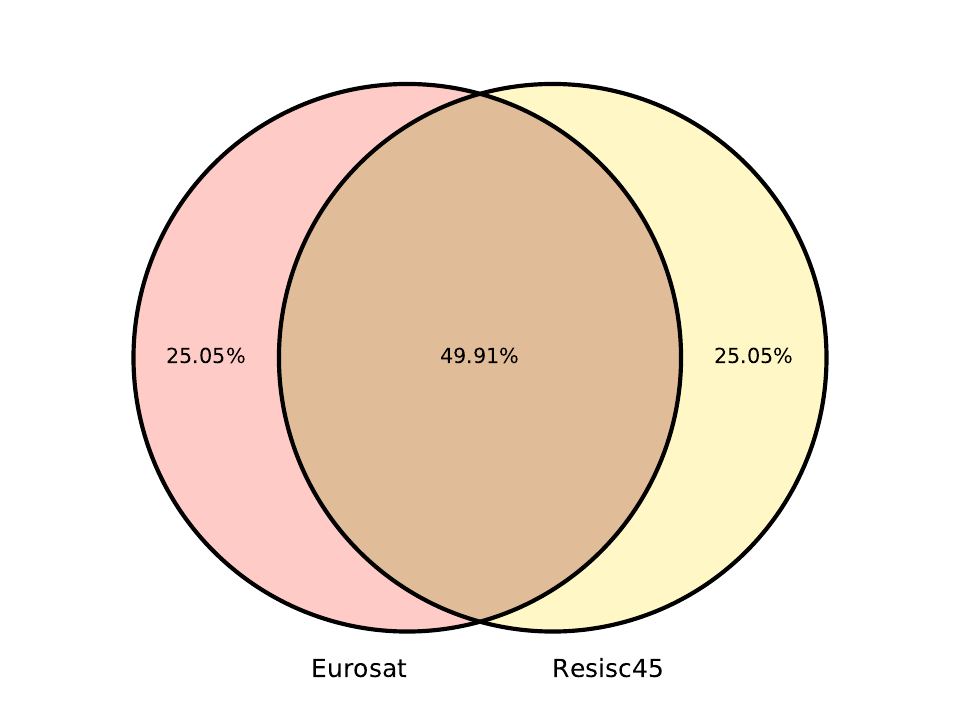}
    }
    \subfigure[]{		\includegraphics[width=0.23\linewidth]{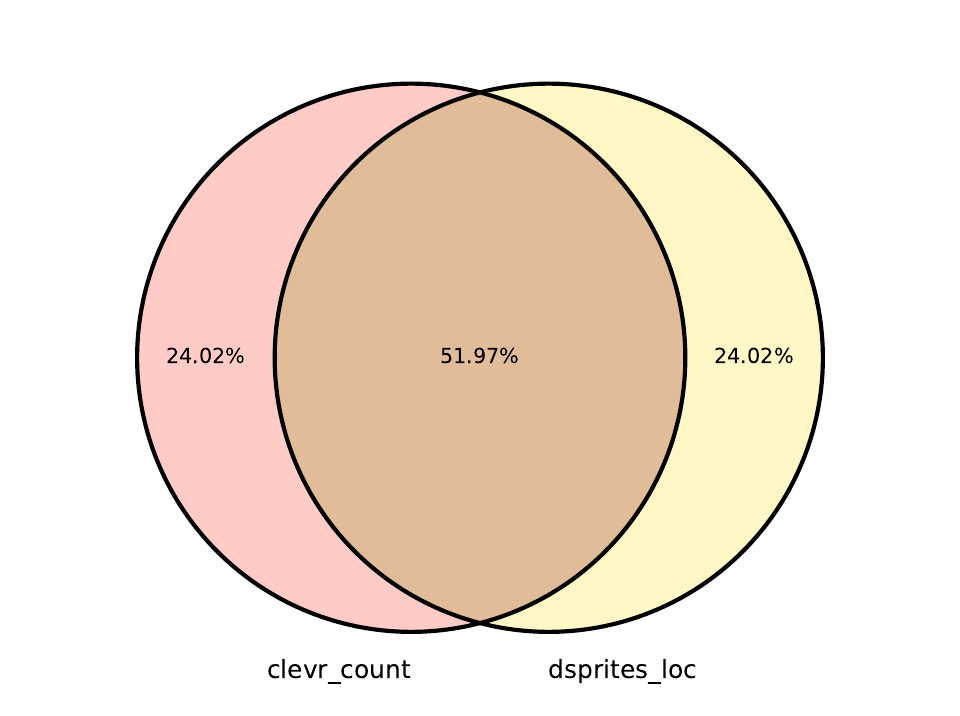}
    }
    \subfigure[]{		\includegraphics[width=0.23\linewidth]{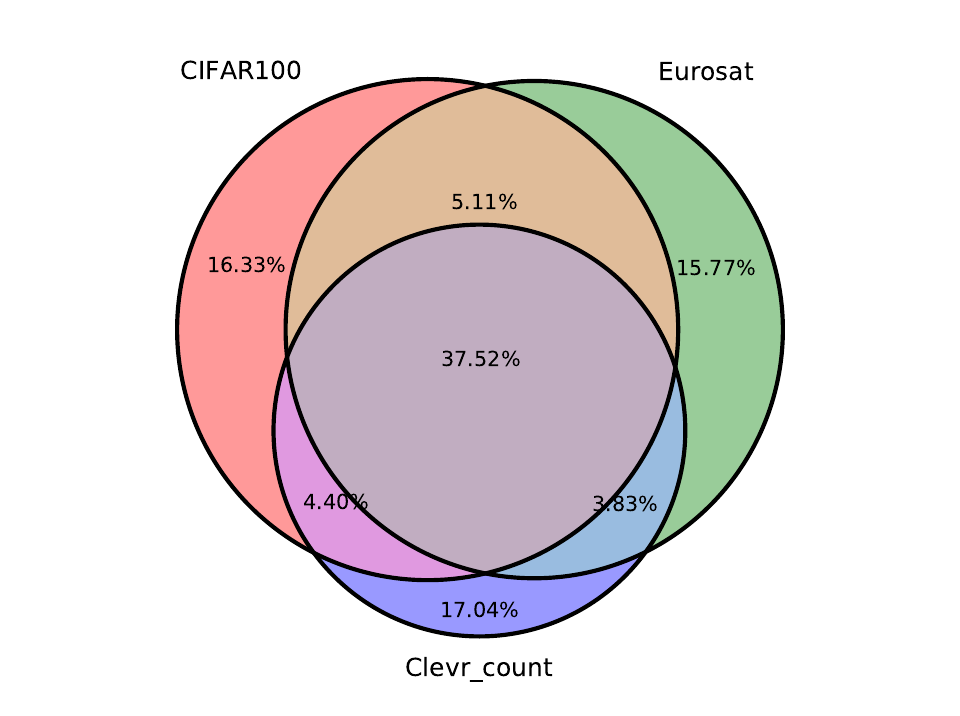}
    }

    \caption{The overlap of the selected parameters across different tasks. Overlap is determined based on parameter position. If selected parameters share the same position in the network, they are considered to have overlap. We test on the ViT-B/16 with following tasks: (a) Cifar100 and Caltech101; (b) Eurosat and Resisc45; (c) Clevr/count and Dsprites/loc; (d) Cifar100, Eurosat and Clevr/count.}
    \label{fig: overlap of different task}
\end{figure*}

\begin{figure*}[!t]
    \centering
    \subfigure[Linear Probing]{		\includegraphics[width=0.23\linewidth]{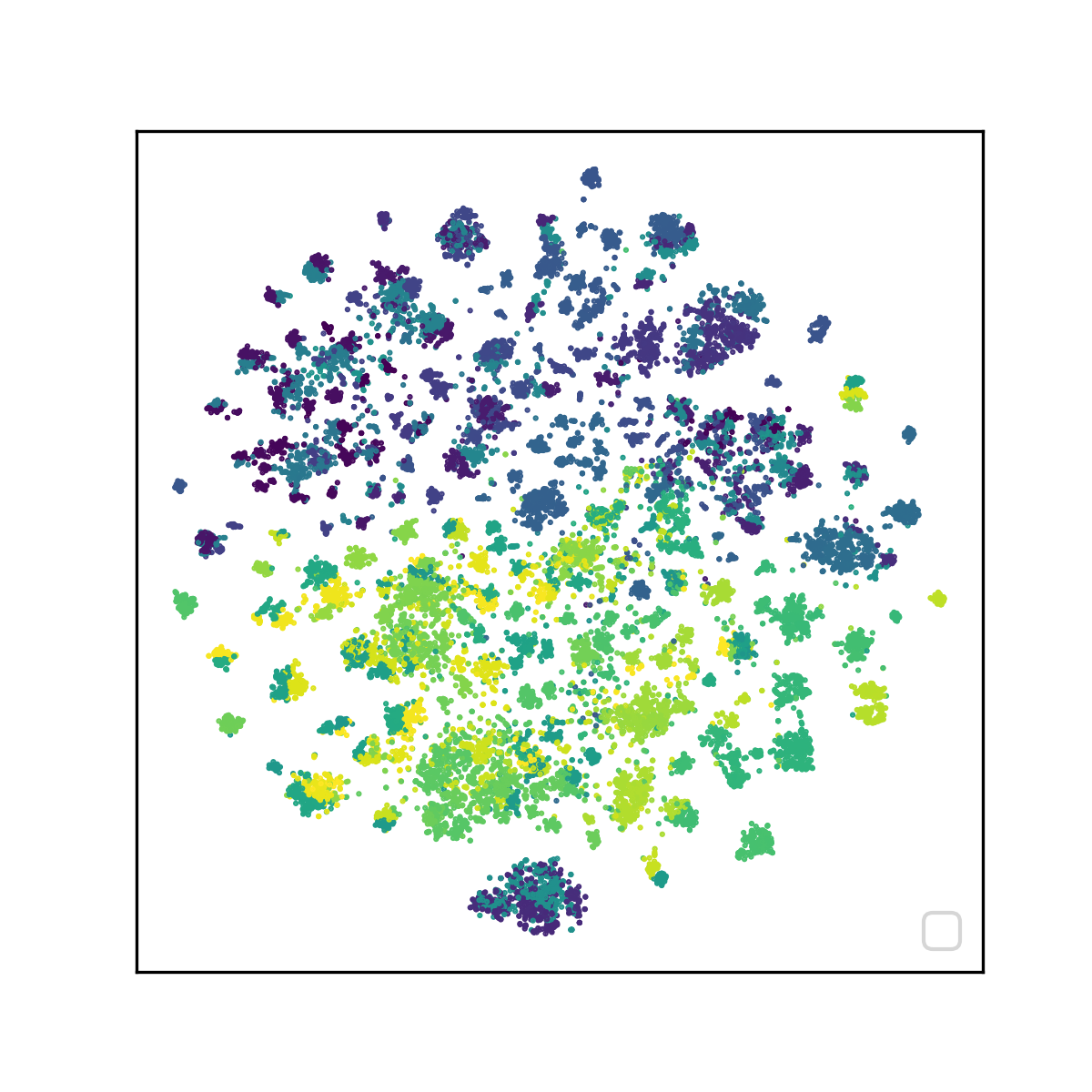}
    }    
    \subfigure[SSF]{		\includegraphics[width=0.23\linewidth]{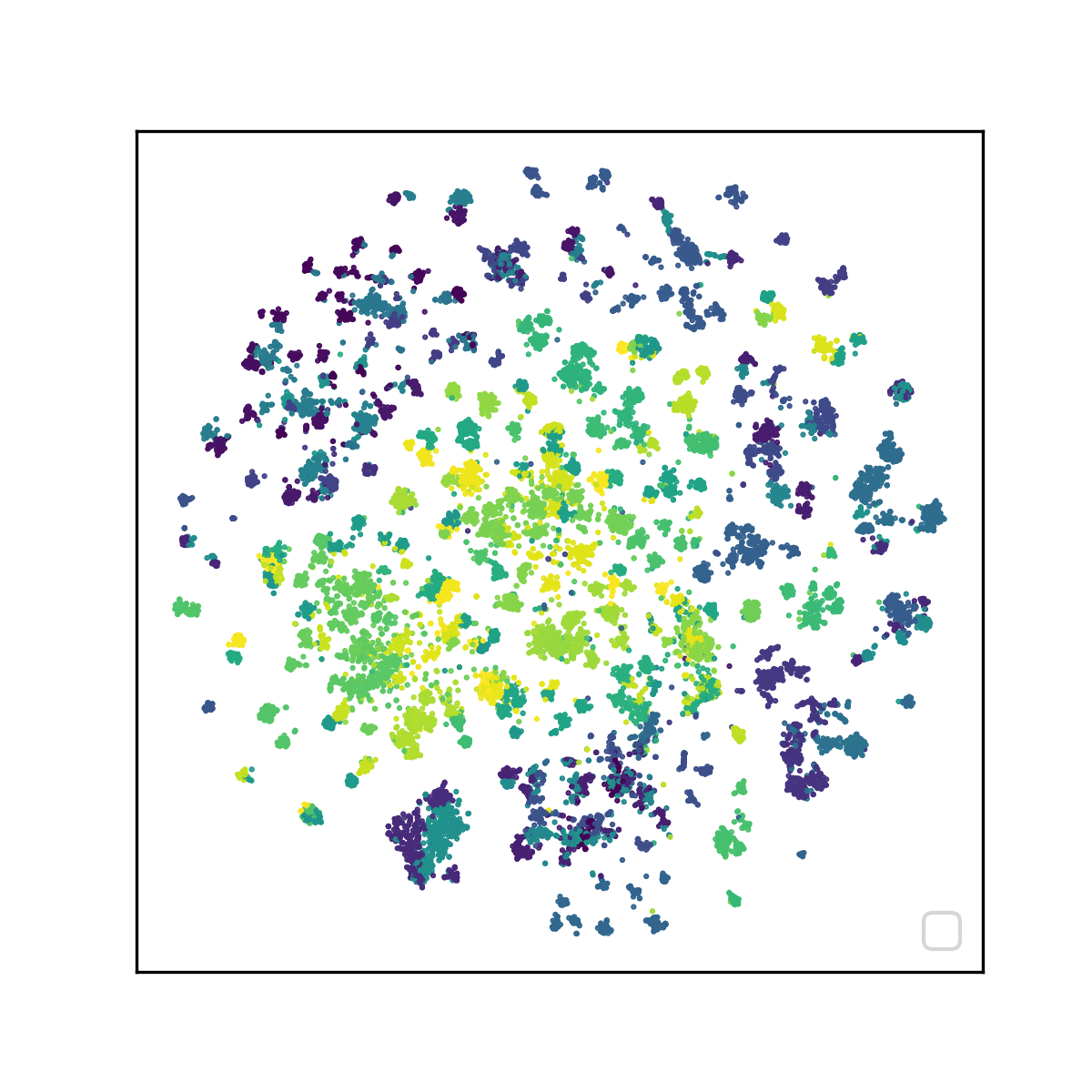}
    }    
    \subfigure[GPS (ours)]{		\includegraphics[width=0.23\linewidth]{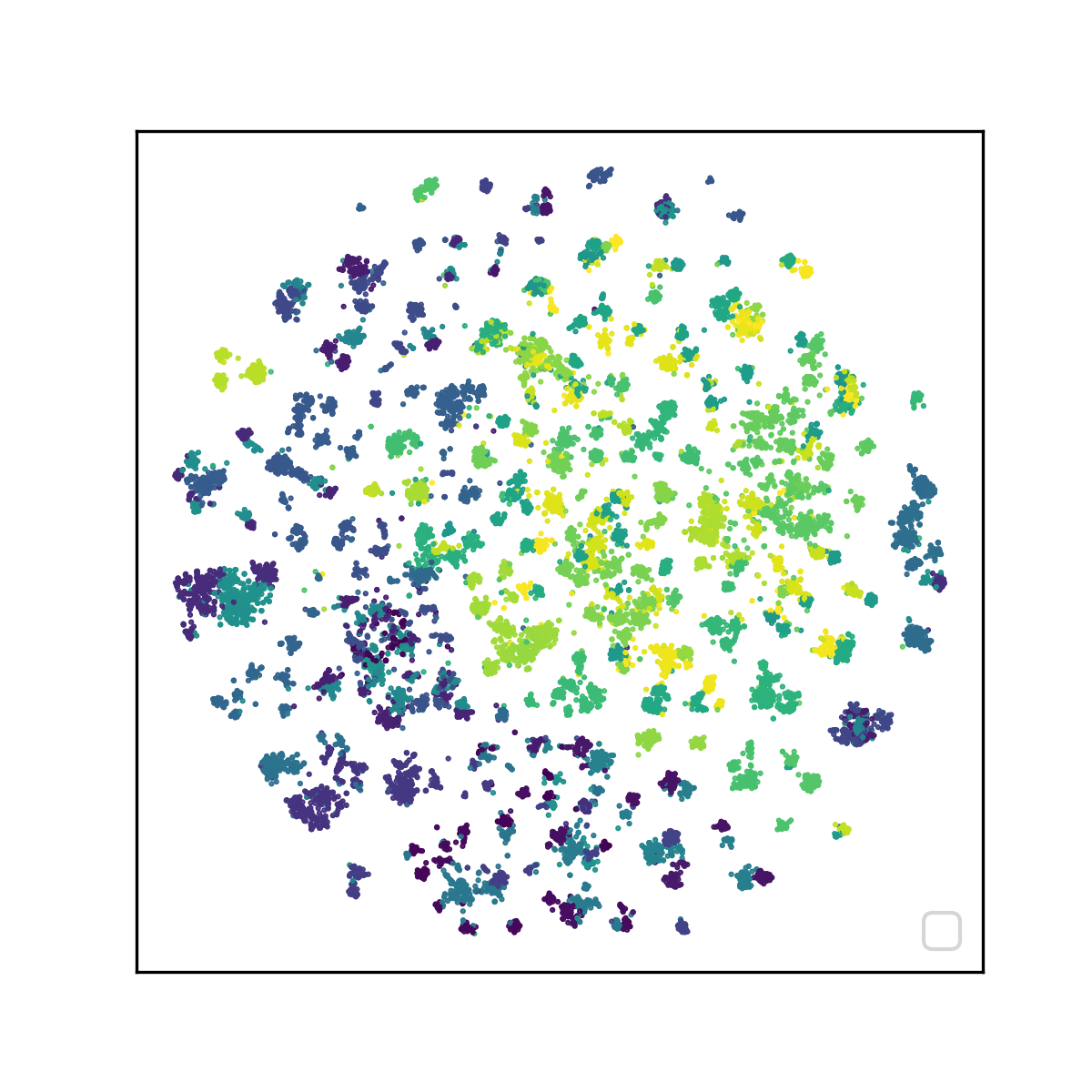}
    }    
    \subfigure[Full fine-tuning]{		\includegraphics[width=0.23\linewidth]{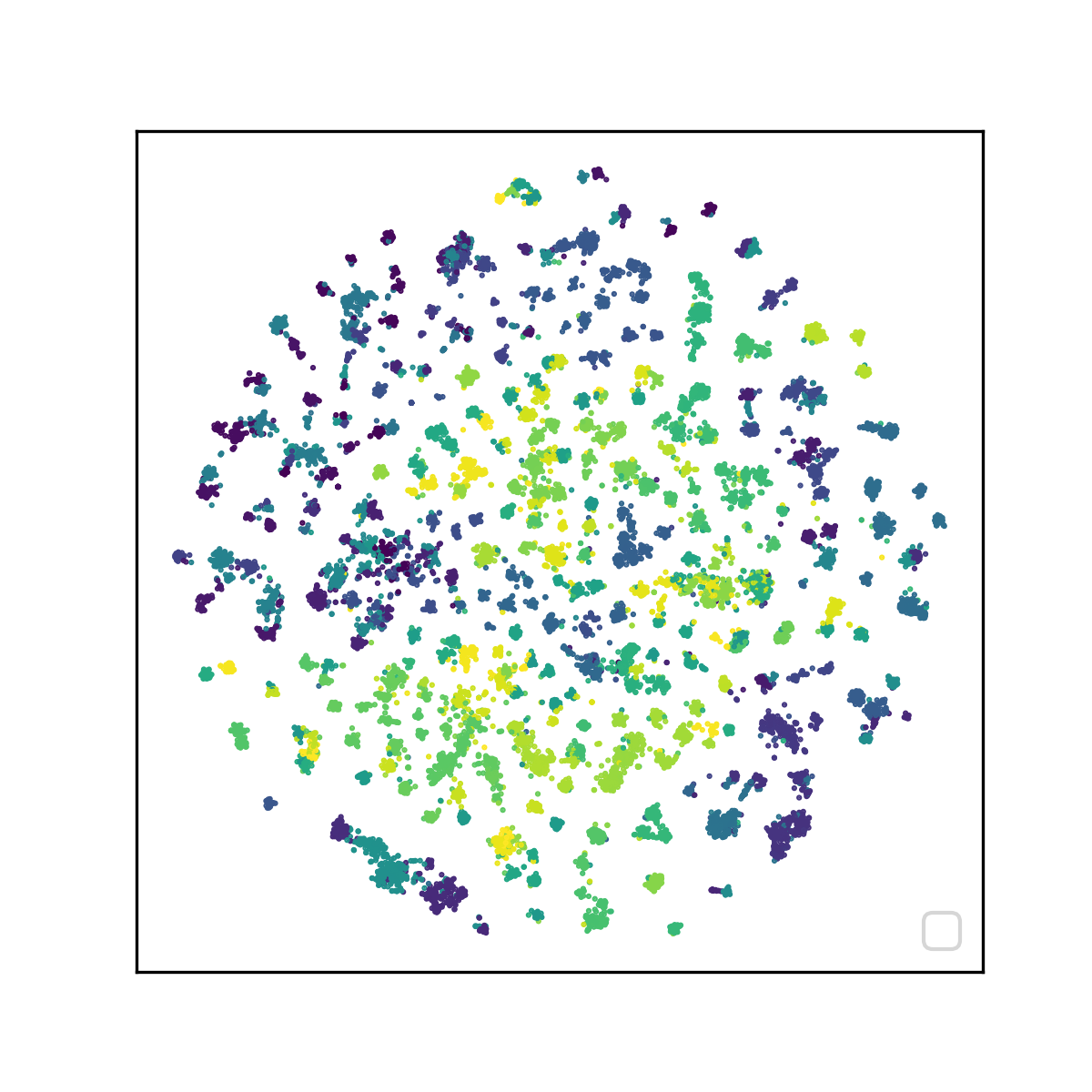}
    }
    \caption{t-SNE visualization of different fine-tuning methods, including linear probing, SSF, GPS, full fine-tuning.}
    \label{fig: tsne}
\end{figure*}

\section{Visualizations}
\label{Visualizations}

\subsection{Semantic segmentation}
\label{Semantic Segmentation}

        
        

As mentioned in the main body of our paper, our approach demonstrates highly promising results in the field of semantic segmentation. We apply our method on the pre-trained strong segmentation model (SAM)~\cite{kirillov2023segany} and fine-tune on a medical segmentation task $-$ polyp segmentation~\cite{jha2020kvasirseg}. Here, we present more case visualizations, which could directly show the effectiveness of our method, as shown in \cref{fig:polyp}.

\subsection{Distribution of selected parameters across various tasks}
\label{Distribution of selected parameters across various tasks}

As we select different subsets of parameters from the original model for different downstream tasks, a normal question is how different the distribution of the selected parameters is across different tasks. we test on Vit-B/16 with 6 downstream tasks from VTAB (two from Natural, two from Specialized and the other two from Structured). As shown in~\cref{fig: overlap of different task}, the chosen parameters exhibit a tendency towards 2/3 shared parameters and 1/3 task-specific parameters, despite the dissimilar data distribution of downstream tasks. This is due to our selection scheme, which makes the parameters evenly distribute  on the whole network and thus the parameters from shallow layers tend to share parameters as similar findings from the field of multi-task learning~\cite{pfeiffer2020adapterfusion, karimi2021compacter, sung2022vl}.

\subsection{Feature distribution}
\label{Feature distribution}

On the NABirds datasets, we use t-SNE to visualize the feature distribution of different fine-tuning methods. The results of all comparison methods are obtained based on the ViT-B/16 pre-trained on ImageNet-21k. The visualization results are illustrated in~\cref{fig: tsne}. Feature clustering results using our GPS are superior to those with linear probing, SSF, and full fine-tuning.

\section{Details of the evaluation datasets}
\label{Details of the evaluation datasets}

\begin{table*}[!t]
\centering
\resizebox{0.85\linewidth}{!}{
\begin{tabular}{llllll}
\toprule
Dataset              & Description                                   & \#Classes & Train                      & Val                   & Test      \\ \midrule
\multicolumn{6}{c}{Fine-Grained Visual Classification (FGVC)}                                                                                     \\ \midrule
CUB-200-2011~\cite{wahcaltech}         & Bird recognition                      & 200       & 5,394                    & 600                  & 5,794     \\
NABirds~\cite{van2015building}             & Bird recognition                      & 555       & 21,536                    & 2,393                & 24,633    \\
Oxford Flowers~\cite{nilsback2008automated}      & Flower recognition                    & 102       & 1,020                      & 1,020                 & 6,149     \\
Stanford Dogs~\cite{khosla2011novel}       & Dog recognition                       & 120       & 10,800                    & 1,200                & 8,580     \\
Stanford Cars~\cite{gebru2017fine}       & Car classification                            & 196       & 7,329                     & 815                  & 8,041     \\ \midrule
\multicolumn{6}{c}{Visual Task Adaptation Benchmark 
(VTAB-1k)~\cite{zhai2019large}}                                                                                    \\ \midrule
CIFAR-100~\cite{krizhevsky2009learning}
           & \multirow{7}{*}{Natural}                      & 100       & \multirow{7}{*}{800/1000}  & \multirow{7}{*}{200}  & 10,000    \\
Caltech101~\cite{fei2006one}          &                                               & 102       &                            &                       & 6,084     \\
DTD ~\cite{cimpoi2014describing}                &                                               & 47        &                            &                       & 1,880     \\
Flowers102~\cite{nilsback2008automated}          &                                               & 102       &                            &                       & 6,149     \\
Pets ~\cite{parkhi2012cats}               &                                               & 37        &                            &                       & 3,669     \\
SVHN~\cite{netzer2011reading}                &                                               & 10        &                            &                       & 26,032    \\
Sun397~\cite{xiao2010sun}              &                                               & 397       &                            &                       & 21,750    \\ \midrule
Patch Camelyon~\cite{veeling2018rotation}       & \multirow{4}{*}{Specialized}                  & 2         & \multirow{4}{*}{800/1000}  & \multirow{4}{*}{200}  & 32,768    \\
EuroSAT~\cite{helber2019eurosat}             &                                               & 10        &                            &                       & 5,400     \\
Resisc45~\cite{cheng2017remote}            &                                               & 45        &                            &                       & 6,300     \\
Retinopathy~\cite{graham2015kaggle}         &                                               & 5         &                            &                       & 42,670    \\ \midrule
Clevr/count~\cite{johnson2017clevr}         & \multirow{8}{*}{Structured}                   & 8         & \multirow{8}{*}{800/1000}  & \multirow{8}{*}{200}  & 15,000    \\
Clevr/distance~\cite{johnson2017clevr}      &                                               & 6         &                            &                       & 15,000    \\
DMLab ~\cite{beattie2016deepmind}              &                                               & 6         &                            &                       & 22,735    \\
KITTI/distance~\cite{geiger2013vision}       &                                               & 4         &                            &                       & 711       \\
dSprites/location~\cite{matthey2017dsprites}   &                                               & 16        &                            &                       & 73,728    \\
dSprites/orientation~\cite{matthey2017dsprites} &                                               & 16        &                            &                       & 73,728    \\
SmallNORB/azimuth~\cite{lecun2004learning}   &                                               & 18        &                            &                       & 12,150    \\
SmallNORB/elevation~\cite{lecun2004learning} &                                               & 9         &                            &                       & 12,150    \\ \midrule
\multicolumn{6}{c}{General Image Classification Datasets}                                                                                         \\ \midrule
CIFAR-100~\cite{krizhevsky2009learning}           & \multirow{2}{*}{General images} & 100       & 50,000                     & -                     & 10,000    \\
ImageNet-1K~\cite{deng2009imagenet}         &                                               & 1,000     & 1,281,167                  & 50,000                & 150,000   \\ \midrule
\multicolumn{6}{c}{Robustness and Out-of-Distribution Datasets}                                                                                    \\ \midrule
ImageNet-A~\cite{hendrycks2021natural}         & \multirow{3}{*}{Robustness \& OOD}            & 200       &         -                   &  -                     & 7,500     \\
ImageNet-R~\cite{hendrycks2021many}          &                                               & 200       &-                            &-                       & 30,000    \\
ImageNet-C~\cite{hendrycks2019benchmarking}          &                                               & 1,000     &-                            &-                       & 75×50,000 \\ \midrule
\multicolumn{6}{c}{Cross-domain Semantic Segmentation Dataset}                                                                                    \\ \midrule
Kvasir-SEG~\cite{jha2020kvasirseg}          &    Polyp Segmentation                               & 2     & 880                           &-                       & 120 \\ \bottomrule
\end{tabular}}
\caption{Detailed statistics of the datasets evaluated on our work. We follow the VPT~\cite{jia2022visual} for train/val split. This table is partially borrowed from VPT\cite{jia2022visual} and SSF~\cite{lian2022scaling}.}
\label{table: dataset details}
\end{table*}

The statistic of all datasets used in this paper is shown in~\cref{table: dataset details}.

\subsection{Image classification}
\label{Image classification}
\paragraph{FGVC}Fine-Grained Visual Classification (FGVC) benchmark includes 5 downstream tasks, which are CUB-200-2011~\cite{wahcaltech}, NABirds~\cite{van2015building}, Oxford Flowers~\cite{nilsback2008automated}, Stanford Dogs~\cite{khosla2011novel} and Stanford Cars~\cite{gebru2017fine}. Each one contains more than 100 classes and a few thousand images. We directly use the public splits if one contains, otherwise, we follow the splits in~\cite{jia2022visual}.

\paragraph{VTAB-1k}Visual Task Adaptation Benchmark~\cite{zhai2019large} contains 19 visual classification tasks which are grouped into 3 sets: (1) Natural $-$ tasks with natural images captured by standard cameras; (2) Specialized $-$ tasks with images captured via specialized equipment, e.g., medical camera or satellite sensor; (3) Structured $-$ tasks with images synthesized from simulated environments, which require geometric comprehension like object counting and depth estimation. Each one contains only 1000 training examples while a large number of test images (i.e. over 20,000 on average).

\paragraph{CIFAR-100}CIFAR-100~\cite{krizhevsky2009learning} is a widely used general image classification task. It contains 50,000 training and 10,000 test images with 100 categories.

\paragraph{ImageNet-1K}ImageNet-1K~\cite{deng2009imagenet} is the most commonly utilized subset of ImageNet for object classification, encompassing 1000 classes and featuring a training set of 1,281,167 images, a validation set of 50,000 images, and a test set of 100,000 images.

\subsection{Semantic segmentation}
\paragraph{Polyp segmentation}
We select kvasir-SEG~\cite{jha2020kvasir} for polyp segmentation task. We follow the settings in Medico automatic polyp segmentation· task at mediaeval 2020~\cite{jha2020kvasirseg} with a train-valid ratio of 880:120.

\subsection{Robustness and OOD}
\label{Robustness and OOD}

\paragraph{ImageNet-A} ImageNet-A~\cite{hendrycks2021natural} contains 200 classes, which is selected from ImageNet-1K (1000 classes). All samples are real-world adversarial samples that caused the ResNet model to produce erroneous classifications.

\paragraph{ImageNet-R} ImageNet-R~\cite{hendrycks2021many} contains art, graffiti, sculptures, tattoos, toys, cartoons, paintings, embroidery, deviantart, graphics, patterns, plastic objects, origami, plush objects,  sketches, and video game renditions from ImageNet classes.

\paragraph{ImageNet-C} ImageNet-C~\cite{hendrycks2019benchmarking} is an open-source collection of algorithmically generated corruptions, such as blur and noise, that have been applied to the ImageNet test set.


\section{Extended related work}
\label{Realted work}
\subsection{Visual parameter efficient fine-tuning}
\label{Visual parameter efficient fine-tuning}
In the field of computer vision, current work endeavors to pre-train larger models~\cite{dosovitskiy2020image, liu2021swin, devlin2018bert,radford2021learning, zhou2021deepvit,zhou2022understanding} on extensive datasets, followed by fine-tuning diverse downstream tasks to achieve superior performance and faster convergence. Conventional arts set all the network parameters learnable and adapt them to the target tasks. However, as foundation models become increasingly large and the number of downstream tasks increases, it becomes impractical due to the significant computational and storage requirements that it entails. Parameter-efficient fine-tuning (PEFT) methods are proposed to alleviate such a burden, which tunes only a tiny portion of the parameters. The general PEFT can be categorized into addition-based and selection-based methods.

Addition-based methods introduce additional parameters to the pre-trained backbone. Adapter methods keep most of the parameters in the model frozen and update only small-scale injected parameters. Bottleneck-structured adapters~\cite{houlsby2019parameter,rebuffi2017learning,bapna2019simple,pfeiffer2020adapterhub,stickland2019bert,pfeiffer2020adapterfusion,ruckle2020adapterdrop,sung2022vl,zhang2021tip,wang2020k} adopt a residual pathway to leverage both original and task-specific knowledge by learning down-projection and up-projection with a nonlinear activation. Others~\cite{mahabadi2021parameter} propose a hyper-network to generate model weights or decompose the dense weighted matrix into the low-rank matrix to reduce parameters~\cite{karimi2021compacter}. Instead of introducing extra modules, prompt methods~\cite{ju2022prompting, gao2020making, hu2021knowledgeable, liu2023pre, ding2021openprompt,li2021prefix,liu2022p} wrap the input with context. A representative work VPT~\cite{jia2022visual} prepend learnable prompts to the input tokens before feeding it into each Transformer block. VPT includes two variants VPT-Shallow and VPT-Deep associated with the number of inserted layers. VPT-Shallow simply prepends prompts to the first transformer layer while VPT-Deep prepends prompts to all the layers. However, it’s inflexible when applying the method to new tasks since it relies on hand-crafted prompt length selection. Apart from the adapter and prompt tuning, a recent study SSF~\cite{lian2022scaling} introduces two learnable vectors to scale and shift the feature map in each transformer operation and achieves promising results. These extra parameters will lead to a substantial increase in computational overhead and hinder the rate of convergence. Our method solves these issues without adding parameters or changing the network topology so it can effectively alleviate such problems.


Selection-based methods~\cite{zaken2021bitfit,guo2020parameter,zhao2020masking} do not introduce any new parameters but directly select part of the parameters to be optimized without modifying the intrinsic architecture of the model. Bitfit~\cite{zaken2021bitfit} only fine-tunes bias vectors in the pre-trained model. Other methods only fine-tune the top-K layers~\cite{houlsby2019parameter} or the last linear layer~\cite{jia2022visual} with other layers freeze. Despite efficiency, they suffer a significant accuracy drop compared to the full fine-tuning since the manually specified parameters tend to be a non-optimal solution. Our gradient-based parameter selection method falls into this category. Since the gradient can serve as a tool for determining parameter significance, our method is intuitive but surprisingly effective.

\subsection{Subset network training}
\label{Subset network training}

Standard pruning technique~\cite{han2015learning,gale2019state,han2015deep,kruschke1991benefits,li2016pruning,wen2016learning}  naturally uncovers subnetworks whose initializations made them capable of training effectively. The lottery ticket hypothesis~\cite{frankle2018lottery} articulate that subnetworks can reach test accuracy comparable to the original network. Drawing from the theory, fine-tuning methods based on subset network are widely studied. SpotTune~\cite{guo2019spottune} designs a policy network to make routing decisions for subset networks. Child-tuning~\cite{xu2021raise} iteratively updates a subset of parameters by masking out the gradients of the non-child network during the backward process. However, the computing overhead led by hyper networks or iterative parameter selection makes none of these methods parameter-efficient. We fix the position of parameters that will be updated by simple gradient weights sorting before training which makes our method parameter efficient.

\section{Disscussion}
\label{Disscussion}
\paragraph{Why sub-network?} There is a lot of research in the field of neural network pruning, where researchers aim to identify the importance of the parameters in a network and eliminate some unnecessary parameters without performance deterioration (about 90\% parameters of the model are pruned)~\cite{1990Optimal, Li2016PruningFF, Chen2020TheLT, Prasanna2020WhenBP}. Motivated by this, we posit the existence of a sub-network containing crucial parameters that can be fine-tuned for optimal performance on downstream tasks.

\paragraph{Magnitude or gradient?} In contrast to the approaches of identifying the importance of the parameters in~\cite{1990Optimal, Li2016PruningFF, Chen2020TheLT, Prasanna2020WhenBP}, which rely on weight magnitude to determine parameter importance, our method identifies parameter importance based on gradient values. An important difference between gradient and magnitude is that the gradient-based method is task-specific, as the gradient is calculated by the backpropagation of the loss for a specific task, while the magnitude-based method use a set of same parameters for all downstream tasks. However, our ablation study in the main body has shown gradient-based method perform bwtter and the~\cref{fig: overlap of different task} also show that each task has it own task-specific parameters.

\section{Limitations and societal impacts}
\label{Limitations and societal impacts}
\paragraph{Limitations}  Several studies~\cite{pfeiffer2020adapterfusion, karimi2021compacter, sung2022vl}  have demonstrated that certain similar tasks can be optimized together through parameter sharing, resulting in improved performance across all individual tasks. However, our work focuses on selecting distinct parameters for various tasks. Although we already tune affordable parameters, we do not fully exploit the potential of parameter sharing across different tasks. Therefore, we posit that our work can be extended to a multitask setting, where tasks share tuning parameters and thus further reduce the total number of learnable parameters.
\paragraph{Societal impacts} Our method can effectively fine-tune pre-trained models for downstream tasks by adjusting less than 1\% of the network's parameters. This is particularly beneficial when dealing with large pre-trained models and multiple downstream tasks, as it saves computational resources, memory costs, and reduces carbon emissions. Our approach maintains the model's original structure without introducing any additional parameters during both the training and inference stages, distinguishing it from other methods. However, similar to other fine-tuning approaches, our method relies on a pre-trained model. If this upstream pre-trained model is trained on illicit data, it may also violate the use of fine-tuning methods.



\end{document}